\def\year{2021}\relax
\documentclass[letterpaper]{article} 
\usepackage{aaai19}  
\usepackage{times}  
\usepackage{helvet} 
\usepackage{courier}  
\usepackage[hyphens]{url}  
\usepackage{graphicx} 
\usepackage{mathtools}
\usepackage{graphicx}  
\usepackage[english]{babel}
\usepackage[utf8]{inputenc}
\usepackage{csquotes}
\usepackage{color,soul}
\usepackage{algorithm}
\usepackage{algpseudocode}
\usepackage{comment}
\usepackage{booktabs}
\usepackage{multirow}
\usepackage[backend=biber,style=numeric,sorting=none, maxbibnames=99]{biblatex}
\usepackage{graphicx}
\pdfoutput=1
\DeclareUnicodeCharacter{0301}{\'{e}}
\usepackage[T1]{fontenc}

\title{Rule Extraction in Unsupervised Anomaly Detection for Model Explainability: Application to OneClass SVM}
\author{Alberto Barbado\textsuperscript{\rm 1,2}, Oscar Corcho\textsuperscript{\rm 2}, Richard Benjamins\textsuperscript{\rm 1}\\
alberto.barbadogonzalez@telefonica.com, ocorcho@fi.upm.es, richard.benjamins@telefonica.com\\ 
\textsuperscript{\rm 1} Telefónica, 28050 Madrid, Spain\\ 
\textsuperscript{\rm 2} Departamento de Inteligencia Artificial, Universidad Politécnica de Madrid, 28660 Boadilla del Monte, Spain\\ 
}
 
\begin{document}

\maketitle

\begin{abstract}
OneClass SVM is a popular method for unsupervised anomaly detection. As many other methods, it suffers from the \textit{black box} problem: it is difficult to justify, in an intuitive and simple manner, why the decision frontier is identifying data points as anomalous or non anomalous. Such type of problem is being widely addressed for supervised models. However, it is still an uncharted area for unsupervised learning. In this paper, we evaluate several rule extraction techniques over OneClass SVM models, as well as present alternative designs for some of those algorithms. Together with that, we propose algorithms to compute metrics related with eXplainable Artificial Intelligence (XAI) regarding the "comprehensibility", "representativeness", "stability" and "diversity" of the extracted rules. We evaluate our proposals with different datasets, including real-world data coming from industry. With this, our proposal contributes to extend XAI techniques to unsupervised machine learning models.

\end{abstract}
\textbf{Keywords} XAI, OneClass SVM, unsupervised learning, rule extraction, anomaly detection, metrics

\section{1. Introduction}

\noindent Responsible Artificial Intelligence (RAI) is defined as the different AI principles that should be considered when developing and deploying real applications based on AI \cite{alej2019explainable}. RAI serves as a methodological framework to both identify core aspects (or AI principles) that should be considered when developing AI solutions while also proposing how to implement them. These AI principles include aspects such as Fairness, Explainability, Security, Privacy and Human-centric design \cite{benjamins2019responsible}.

The AI principle of Explainability is addressed through the use of Explainable AI (XAI) techniques, which can be applied to black-box models in order to obtain post-hoc explanations based on the information that they provide. In the literature, there are many XAI proposals for supervised ML models. However, some of the most recent and thorough reviews on XAI \cite{gilpin2018explaining, mueller2019explanation, alej2019explainable, molnar2019interpretable} do not mention many applications of those techniques to unsupervised learning.

Outlier detection is one of the tasks where unsupervised learning is applied. It is defined as the process of detecting anomalous observations within a dataset, and sometimes remove it as a first step within data-mining applications \cite{hodge2004survey}. There is often no prior information about outliers in a dataset, hence unsupervised ML algorithms offer the chance to infer patterns and detect anomalies. However, not only is it important to detect outliers, but also to explain
them. Explanations can help to understand why a particular datapoint has been labelled anomalous (and what changes in the feature values would lead to classify it as an inlier), and how the model behaves globally (for instance, what features influence more for classifying a datapoint as an outlier).

The output of an unsupervised ML model for anomaly detection can be seen as binary (an observation may be an "outlier" or an "inlier). Thus, surrogate post-hoc XAI techniques can yield explanations similarly to a supervised binary classifier where the two possible outputs are imbalanced. Hence, the explanations for the model can be obtained by using XAI techniques already designed for supervised ML binary classifiers. This is already addressed in the literature, particularly by using feature-relevance XAI techniques \cite{ruff2021unifying, langone2020interpretable}.

Among the different model-agnostic post-hoc XAI techniques that can be applied, rule extraction offers the possibility to provide both global and local explanations, as indicated by the recent literature \cite{alej2019explainable}. This is achieved by using an "IF...THEN" schema that explains both the output of a particular datapoint as well as the global behaviour of the original model. In the case of outlier detection, they can explain both a particular outlier and also how the features of the whole model contribute to identify points as outliers or inliers. Even though there are some examples of this in the literature, particularly for the case of OCSVM \cite{padmaja2015hybrid}, there are not many studies covering it to the best of our knowledge.

There is a particularity of the usage of rule extraction for explaining anomalies. An outlier detection system that uses rules as explanations may have more interest in explaining faithfully why a datapoint is an outlier, and what should have happened in order for it to be an inlier, rather than being able to cover all possible scenarios with explanations that may be wrong. This means that the extracted rules need to have a 100\% precision (P@1); rules that classify datapoints from one class (i.e. "outliers") without including datapoints from the other one. Considering the example of rules extracted that cover inliers, this is important because the counterfactual explanation for how to turn an outlier into an inlier should lead to a scenario where the model will always classify it as an inlier.

This is linked to another aspect regarding XAI. Even though there are many model-agnostic post-hoc rule extraction techniques that can be used for explaining a ML model in general, and an unsupervised one for anomaly detection in particular, there is still one question present: which technique provides the best explanations?. This leads to an open issue within the XAI literature: how to evaluate the quality of explanations?. Here, the literature suggests some concepts to consider while designing new metrics and algorithms. The metrics need to consider the type of explanations provided (rule based in this case) and the type of data used. For instance, in our case we work on anomaly detection without a prior ground truth. Hence, some XAI metrics (like those related to accuracy measurement of the rule predicitons over a test set) are not applicable. Together with that, other particularities of the problems addressed may influence in what metrics are more important. In our case, since we are using P@1 rules, measuring the fidelity of the explanations is not necessary (since the comparison will only be possible against the model output). However, other metrics gain more relevance, such as stability. With that, some relevants aspects to measure for this case are: 

\begin{itemize}
\item "Comprehensibility": Are explanations easy enough to understand?
\item "Representativeness": Are explanations relevant? Do they explain all possible cases?
\item "Stability": Do explanations match the predictions of the model? Or are there inconsistencies?
\item "Diversity": Are explanations sufficiently different among them? Or are they redundant?
\end{itemize}
There are only a few implementations of these concepts as metrics to measure and compare rule extraction techniques (and even though, mainly for rule extraction techniques over supervised ML).

Following this, the main contributions of our work are:
\begin{itemize}
\item Applying different model-agnostic rule extraction techniques to explain the anomalies detected by an unsupervised OCSVM ML model through P@1 rules and for different types of kernels (RBF and Linear). We use existing rule extraction techniques and propose some variations over the one described in
\cite{nunez2002rule}.
\item Quantifying the quality of the generated P@1 rule explanations for unsupervised anomaly detection with OCSVM using XAI metrics that measure comprehensibility, representativeness, stability and diversity. Particularly, we propose novel ways to implement stability and diversity.
\item Evaluating how the comprehensibility aspect regarding the number of P@1 rules generated significantly varies depending on the kernel considered (Linear vs RBF) and depending on whether the explanations are for outliers or for inliers.
\end{itemize}

The empirical evaluations carried out use both open datasets as well as real data from Telefónica. Our evaluation consists in analysing the results for the aforementioned metrics using different rule extraction algorithms over OCSVM models with two kernel configurations, Linear and RBF.

The rest of the paper is organized as follows. First, we describe some related work in the area of XAI and rule extraction applied to SVM. This chapter will also introduce the rule extraction techniques considered, as well as literature related to XAI metrics and the psychology of explanations. After identifying research opportunities derived from these works, the paper introduces some alternatives over some of the rule extraction techniques, as well as algorithms to compute the metrics described before. Following this, we present an empirical evaluation of our algorithm with several datasets. We then conclude, showing also potential future research lines of work.

\section{2. Related Work}
This section reviews unsupervised ML models used for anomaly detection, and reviews previous work on rule extraction in SVM that is relevant for our proposal.

\subsection{2.1. Unsupervised ML for Anomaly Detection}
The review of \cite{ruff2021unifying} provides an extensive analysis of the SOTA of ML models for anomaly detection, including unsupervised ones. Unsupervised ML models for anomaly detection can be differentiated according to their feature map, or according to the type of model used (in terms of how the decision frontier is obtained). Regarding the feature map, there are two possible types. First, Shallow models (i.e. Minimum Volume Ellipsoid) versus Deep ones (i.e. Generative Adversarial Networks). Regarding the type of model, four types are mentioned: classification (i.e. OCSVM), probabilistic (i.e. Kernel Density Estimation), reconstruction (i.e. Principal Component Analysis, Deep AutoEncoders) and distance-based (i.e. IsolationForest, Local Outlier Factor). 
OCSVM is a type of Kernel-based One-Class Classification anomaly detection model that is well-suited for multimodal, nonlinear and nonconvex datasets. OCSVM is also an algorithm that, since its original formulation \cite{scholkopf2000support}, has being developed with many variations.

OCSVM has advantages in terms of computational performance \cite{wang2004anomaly}. One of the reasons is that it creates a decision frontier using only the support vectors (like general supervised SVM). Another advantage is that model training always leads to the same solution because the optimization problem is a convex one. However, SVM (hence OCSVM) algorithms are difficult to explain due to the mathematically-complex method that obtains the decision frontier \cite{alej2019explainable}.

From a theoretical point of view, SVM for classification maps the data points available in the dataset to a higher dimensional space than the one determined by their features, so that the separation among classes may be done linearly. It uses a hyperplane obtained from data points from all of the classes. These data points, known as support vectors, are the ones that are closer to each other and the only ones needed to determine the decision frontier. However, it is not really necessary to map to a higher dimension due to the fact that the equation that appears in the optimization of the algorithm uses a dot product of those mapped points. Because of that, the only thing to be calculated is such dot product, something that can be accomplished with the well-known kernel trick. Hence instead of calculating explicitly the mapping to a higher dimension the equation is solved using a kernel function. 

In OCSVM there are no labels. Hence all data points are considered to belong to a same class at the beginning. The decision frontier is computed trying to separate the region of the hyperspace with a higher number of data points close to each other from another that has small density, considering those points as anomalies. To do so the algorithm tries to define a decision frontier that maximizes the distance to the origin of the hyperspace and that at the same time separates from it the maximum number of data points. This compromise between those factors leads to the optimization of the algorithm and allows obtaining the optimal decision frontier. Those data points that are separated are labeled as non-anomalous (+1) and the others are labeled as anomalous (-1).

The optimization problem is reflected in the following equations:
\begin{equation}\label{eq1}
\begin{split}
  min_{w, \xi_i, \rho} = \frac{1}{2} ||w||^2 + \frac{1}{\nu n}\sum_{i=1}^{n}(\xi_i - \rho) \\
  \text{subject to:}\\
  (w, \phi(x_i)) \geq \rho - \xi_i\:\:for\:i = 1,...,n  \\
  \xi_i \geq 0\:\:for\:i = 1,...,n
\end{split}
\end{equation}

In that equation, $\nu$ is a hyper-parameter known as \textit{rejection rate}, which needs to be selected by the user. It sets an upper bound on the fraction of anomalies that can be considered, and also defines a lower bound on the fraction of support vectors that can be considered. Using Lagrange techniques, the decision frontier obtained is the following one:
\begin{equation} \label{eq2}
\begin{split}
f(x) = sgn((w, \phi(x_i) - \rho) \Rightarrow \\
f(x) = sgn(\sum_{i=i}^{n}\alpha_i K(x_i,x) - \rho)
\end{split}
\end{equation}

Hence the hyper-parameters that must be defined in this method are the rejection rate, $\nu$, and the type of kernel used. 

\subsection{2.2. Rule Extraction techniques in XAI}
Rule extraction belongs to the group of post-hoc XAI techniques \cite{alej2019explainable}. This group of techniques are applied over an already trained ML model (generally a blackbox one) in order to explain the decision frontier inferred by using the input features to obtain the predictions. Rule extraction techniques are further differentiated into two subgroups: model specific and model-agnostic. Model specific techniques generate the rules based on specific information from the trained model, while model-agnostic ones only use the input and output information from the trained model, hence they can be applied to any other model. Post-hoc XAI techniques in general are then differentiated depending on whether they provide local explanations (explanations for a particular data point) or global ones (explanations for the whole model). Most rule extraction techniques have the advantage of providing explanations for both cases at the same time.

\subsubsection{2.2.1 Model Specific Rule Extraction techniques in XAI for SVM} 
\leavevmode\newline
\cite{barakat2010rule} offers a review of rule extraction techniques for SVM, including the ones that are model specific. Here, regarding model specific techniques, they highlight three different types of algorithms. The first of them are rule extraction algorithms that use the support vectors from the original model as an input source for generating the rules. This is the case of SQRex-SVM \cite{barakat2007rule} where the authors propose the usage of a subset of the support vectors for inferring the rules with the usage of a modified sequential covering algorithm. 
The second type of algorithms use both information from the support vectors together with information from the separating hyper-plane. This is the case of RulExSVM \cite{fu2004extracting}, where the authors propose a technique applicable for SVM with a RBF kernel. The algorithm uses the support vectors in order to build hyper-rectangles that intersect with the separating hyper-plane. Finally, the last type of techniques use the support vectors, the separating hyper-plane, and the training data. The training data is used to define the regions in the hyperspace, and the support vectors and the hyper-plane define the size of those regions. Within this category appears the proposal of \cite{nunez2002rule}.

The authors of \cite{nunez2002rule} propose a technique called SVM+ Prototypes that can be considered model-agnostic or model specific depending on how is implemented. The general intuition consists in finding hypercubes (or hyperspheres) using the centroids (or prototypes) of data points of each class. Then, it can use as vertices either the support vectors from the SVM model, or the data points from that hyperspace area farther away from that centroid. For the first alternative, the proposal is model specific, since it focuses on a specific component of the model itself (the support vectors). The second one is model-agnostic, since it does not use any information that is specific only for SVM models.
After this, it infers a rule from the values of the vertices of the hypercube that contain the limits of all the points inside it, creating one rule for each hypercube.

\begin{figure}[h!]
\centering
 \begin{tabular}{c@{\qquad}c@{\qquad}c}
\includegraphics[width=0.72\columnwidth]{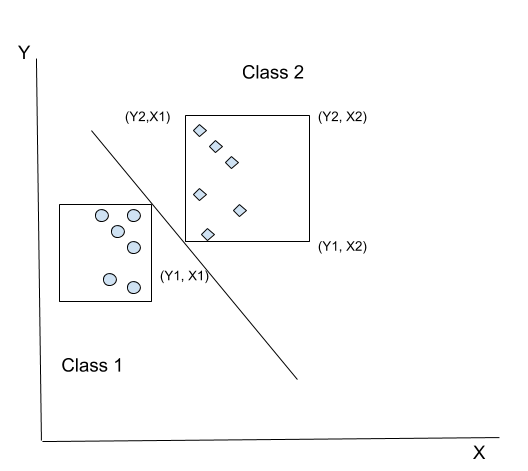}
  \end{tabular} 
  \caption{SVM with linear kernel classifying data points of two classes.\label{fig:outlier0}}
\end{figure}

For example, a dataset that contains two numerical features X and Y will be defined in a 2-dimensional space. The algorithm will create a square that contains the data points on each of the classes, as shown in Figure \ref{fig:outlier0}. The rule that justifies that a data point belongs to class 2 is:
\begin{itemize}
    \item Rule 1: CLASS 2 IF X$\geq$X1 $\land$ Y$\geq$Y1 $\land$ X$\leq$X2 $\land$ Y$\leq$Y2
\end{itemize}

\begin{figure}[h!]
\centering
  \begin{tabular}{c@{\qquad}c@{\qquad}c}
\includegraphics[width=0.70\columnwidth]{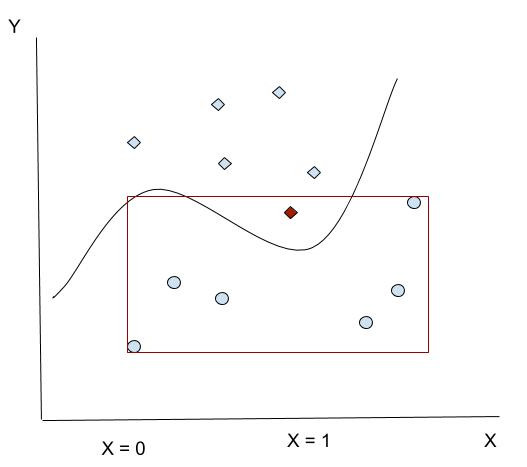}
  \end{tabular} 
  \caption{A hypercube generated using the farthest points leads to the wrong inclusion of data from the another class.\label{fig:outlier1}}
\end{figure}

\begin{figure}[!h]
\centering
  \begin{tabular}{c@{\qquad}c@{\qquad}c}
\includegraphics[width=0.70\columnwidth]{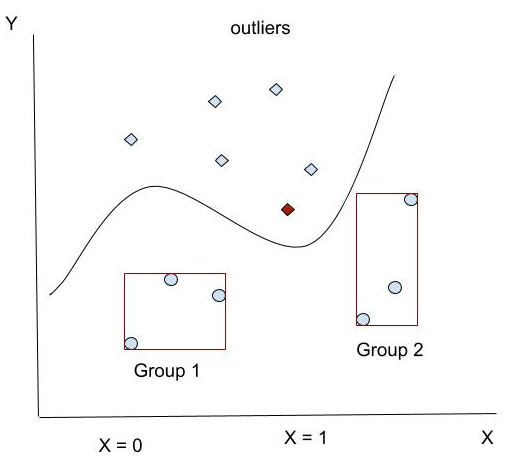}
  \end{tabular} 
  \caption{Using more hypercubes avoids the aforementioned problem. Now there is no wrong inclusion of data points from another class.\label{fig:outlier2}}
\end{figure}

The generated hypercubes may wrongly include points from the other class when the decision frontier is not linear or spherical, as shown in Figure \ref{fig:outlier1}. In this case, the algorithm considers an additional number of clusters trying to include the points into a smaller hypercube, as shown in Figure \ref{fig:outlier2}.

A rule will be generated for each hypercube, considering all those scenarios as independent, leading to this output:
\begin{itemize}
    \setlength{\itemindent}{2em}
    \item Group 1: CLASS 1 IF X...
    \item Group 2: CLASS 1 IF X...
\end{itemize}

There are some downsides of that method in supervised classification tasks, especially when the problem is not simply a binary classification or when the algorithm is performing a regression. For instance, the number of rules may grow immensely due to the fact that a set of rules will be generated for each category and each set may contain a huge number of rule groups, leading to an incomprehensible output.

However, in OCSVM these difficulties may be potentially mitigated due to two reasons. On the one hand, the explanations are reduced to rules that explain when a data point is not an anomaly (so there would be no need to define rules for the anomalies). On the other hand, the algorithm tries to group all non-anomalous points together, setting them apart from the outliers. Because of this, the chance to define a hypercube that does not contain a point from the another class may be higher than in a standard classification task. Both the unbalanced inherent nature of data points in anomaly detection (few anomalies vs. many more non-anomalous data points) and the fact that non-anomalous points tend to be closer to each other may help achieving good results with this method.

\subsubsection{2.2.2 Model-agnostic Rule Extraction techniques in XAI}
\leavevmode\newline
Many rule extraction proposals contribute to XAI without the need to use any specific information from a particular type of model \cite{alej2019explainable}. The only information necessary for building the rules is the input features and the model outputs. Some techniques use all the training data, while others need only a few input instances, or they can even generate artificial datapoints to infer the decision frontier.
Even though the techniques were initially conceived for supervised ML, they can be extended for unsupervised ML for anomaly detection, since there output is analogous to a binary classifier where the classes are heavily imbalanced, as discussed in Section 1.

A general way to approximate any blackbox model globally is by using a surrogate supervised decision model trained over the same dataset, but instead of using the real labels (the ones used for the blackbox model), it is trained over the predictions of that blackbox model \cite{molnar2019interpretable}. This may be accomplished with any ML model, but it is useful to do it with a whitebox model that can be directly interpreted. Among these whitebox models, some of them may be used for rule extraction. An example is a Decision Tree (DT) model. DT allows explaining the classification logic of the blackbox model through the usage of rules, which can be used even for classifying new instances. The advantages of using a DT as a surrogate global model is its flexibility (it can be applied over any model in an agnostic way) and simplicity (it is a solution that is easy to explain). However, this approximation at the end leads to explain a proxy model, and not the actual data, since the surrogate model never sees the true target values.

Anchors \cite{ribeiro2018anchors} is a model-agnostic XAI technique that extracts rule explanations for individual data points. The purpose of Anchors is finding a decision rule that approximates the decision function of the blackbox model around that individual data point. This rule "anchors" the prediction of that data point, so that any perturbation of the features of that point that are still inside the rule will always return the same output from the blackbox model. The approach is as follows. First, the algorithm generates candidate rules that may explain the data point. Then, it evaluates those candidate rules. In order to do that, Anchors generates permutations around the data point (similar data points to the original one) that yield the same result. The result is evaluated by calling the blackbox model (the oracle) and obtaining the classification for that data point. In order to optimize the exploration-exploitation of generating and evaluating data points, it uses a reinforcement learning approach with a Multi-Armed Bandit (MAB) approximation. In this MAB, each arm of the Bandit problem is a candidate rule, and the data points generated, after obtaining their classification result from the blackbox model, are used to compute a precision metric used to evaluate the candidate rule's payoff. This reinforcement learning approach helps minimizing the number of calls to the model in order to reduce the computational cost of the algorithm. Among all the candidate rules, the algorithm then checks if the best one of them matches a predefined convergence criteria. To do that, it filters rules according to a precision threshold, and selects form the remaining ones the one with highest coverage. That rule is used to explain that original data point. If there are no rules that match the convergence criteria, then the algorithm keeps iterating (using a beam search approach) using the B best rules from the previous step in order to generate new candidate rules for the following one. In those following steps, Anchors keep extending the rules with more features (in the first step, it only uses one feature per candidate rule).
Thus, Anchors offers a model-agnostic approach that generate IF-THEN rules, easy to interpret, that are generated in an efficient way thanks to the usage of reinforcement learning (MAB) that can be parallelised. However, Anchors is very sensitive to its initial configuration, like many permutation approach algorithms, such as LIME \cite{ribeiro2016should}. Another important consideration of Anchors is that, while it keeps the calls to the oracle to a minimum (thanks to MAB), it still requires a lot of calls, and that can affect the runtime of the algorithm.

RuleFit \cite{friedman2008predictive} is a model-agnostic surrogate model that learns a linear regression model (Lasso) that uses as features both the original features of the model, as well as new generated features that represent decision rules. In order to accomplish that, first, a tree model is trained over the output and the input features, and the decision paths between the tree levels are turned into decision rules, except for the ones that lead to the leaf nodes, which are not considered. These rules are used as additional features, along with the original ones, on the Lasso surrogate model. Thanks to this, RuleFit yields both rules as well as their contribution, measured through the coefficients of the Lasso model. In summary, RuleFit generates a white-box model that includes rules as features, that can be interpreted as a standard linear regression one. The only caveat is that, for the original coefficients, the predicted outcome changes by $|\beta_j$ if feature $x_j$ changes by one unit if the other features remain unchanged, while for a feature-rule $r_k$ it is different; if all the conditions of the feature $r_k$ are met, the predicted outcome changes by $\alpha_k$ (the weight associated to that rule-coefficient) for regression. Similarly, for classification tasks, when the conditions of $r_k$ are met, the odds for event vs. no-event changes by a factor of $\alpha_k$.

Similarly to RuleFit, SkopeRules \cite{molnar2019interpretable} is another way to generate rules from tree ensembling techniques. They differ, however, in how they obtain the rules. 
First, SkopeRules generates the rules using surrogate tree ensembles trained using the input features and the target variable. Then, it applies a filtering step in which, using a threshold for Precision and Recall, some rules are removed and some are kept. This step allows to select only high-performing rules, and removing the ones that do not yield good results. The last step is known as "semantic rule duplication". This step eliminates duplicate rules (rules that are the same or very similar to other ones). It also eliminates again low-performing rules based on their results for a F1-metric. This allows to obtain high-performing as well as heterogeneous rules. The final set of rules is the output of SkopeRules, differing from RuleFit because it does not use a Lasso model to aggregate all rules.

Falling Rule Lists (FRL) \cite{wang2015falling} are classification models that generate a sorted list of IF-THEN rules, thus, they can serve as a model-agnostic global post-hoc rule extraction technique. The rules are binary, and are looked one after the other, in order to see if a particular datapoint can be classified into one of the classes. The rules are sorted according to the probability of classifying a datapoint into that class using that rule. Due to that, FRL offers a list of IF-ELSE IF rules associated to a particular class with a decreasing probability score. This is inspired in the concept of healthcare triage: patients are classified within risk level groups, and the highest-risk ones should be considered first. The particular algorithm cited and used in this paper uses an approach for learning based on a Bayesian framework, instead of a greedy decision tree learning method, named Bayesian Falling Rule Lists (BFRL). 

Boolean Decision Rules via Column Generation (BRCG) \cite{NIPS2018_7716} also provides a binary classifier by using disjuntive normal form (DNF, OR-of-ANDs) or conjuntive normal form (CNF, AND-of-ORs) through interpretable rules. In case of DNF (the one used in this paper), they provide an unordered set of decision rules that classify a datapoint into the positive category if at least one of the rules is satisfied. This is different than other methods already mentioned, such as BFRL where the rules are ordered in an IF-THEN schema, or the surrogate DT model, that provides the rules in a tree structure schema.
In this article, we use the BRCG-light approximation from \cite{aix360-sept-2019}, that replaces the integer programming solver used in the original paper by a heuristic beam search one.

Generalized Linear Rule Models (GLRM) \cite{pmlr-v97-wei19a} generate decision rules and combine within a linear model (generalized additive model, GAM). Thus, they provide both a non-linear modelling, thanks to the decision rules, while keeping the interpretability by using a linear model that ensembles them. However, as \cite{arya2019one} notice, while it is feasible to interpret linear combinations of rules, if the number of rules increases too much, there is a risk of losing the interpretability of the model. The authors of the original paper highlight that in order to reduce the rules generated and not lose interpretability, they use a rule selection technique based on column generation (CG). CG searches the spaces of rules and generates them only when they are needed, and then fits again the GLM model. This allows analysing again old rules, re-weight them, and discard the ones that are not needed anymore. This is different to other methods used in the literature, mainly pre-selecting a subset of candidate rules using optimization techniques, or a greedy optimization approach by adding rules one by one using sequential covering or boosting techniques.

\subsection{2.3. XAI for Anomaly Detection}
XAI is useful for both explaining an anomaly detection model from a global perspective, or for explaining the identification of particular instances as outliers. From the global explanation level, \cite{tallon2020explainable} use two anomaly detection ML algorithms (Decision tree and DeepLog) to detect outliers over log data. Together with that, they use Shapely values in order to generate model-agnostic feature relevance explanations that help to see which features contribute more for predicting outliers by seeing the individual contribution of each feature to the general outlier probability. 

XAI \cite{langone2020interpretable} has also been used for anomaly detection for predictive maintenance. The authors highlight that even when an anomaly detection model is very accurate, the operators that will get the model prediction may not trust it if it remains a blackbox that does not provide any insights about its decisions. Because of that, they propose an anomaly detection system where the explanations are generated thanks to the usage of a whitebox model (ElasticNet Logistic Regression). So, they provide explanations in terms of feature relevance, focusing on explaining what contributes to an anomalous state. With that, they highlight that explanations for anomaly detection can be generated in a similar way to those of a supervised ML model for binary classification (even though anomaly detection models provide an output heavily imbalanced)

Shapely values for explaining anomalies also appear at \cite{mitani2020highly}, where the SHAP algorithm is used to generate feature relevance explanations in order to explain what contributes specimen mix-up. For the anomaly detection, they use a Gradient Boosting Tree in order to be able to learn efficiently from highly unbalanced data while yielding good predictions. The authors highlight the importance of having a highly accurate model that is able to predict correctly the specimen mix-up, because this is a crucial problem that may lead to an incorrect diagnostic or an inappropriate therapy.

An additional recent reference is \cite{ruff2021unifying}. Here, the authors also cover within the review of the SOTA of anomaly detection the importance of using XAI in order to have a deeper understanding of the model. Their focus on explanations is mainly for unsupervised deep learning (DL) models, where the explanations can be produced by model-agnostic post-hoc techniques for feature relevance (LIME) or by using model specific algorithms (LRP). One of the usages of XAI that they describe is the improvement of the model based on the explanations provided. They show an example for anomaly detection based on images, where XAI helps to see the cases where the pixels used for making the decision are actually the correct ones.

The analysis of the literature highlights how detecting anomalies is critical within some domains, and because of that, their detection needs to be very precise. However, being able to detect anomalies is not enough, and explanations are needed for both understanding the model better (and seeing if it can be trusted or improved), as well as for explaining the model for other audiences in order to see if they can also rely on the predictions or not (something connected to the explanation generation for different user profiles \cite{alej2019explainable}). A model may perform apparently very well and explanations may help to see that the model is taking its decision by using features that are not relevant \cite{molnar2019interpretable}, so in that case, the model may not be finally trusted. This shows that XAI can complement the classical evaluation of models based only on their performance.

However, after the assessment of a model and seeing that it behaves correctly (from both the XAI and the performance point of view), before providing explanations to some user profiles, it is important to ensure that they are aligned to what the model predicts, and are not showing any contradictory information. A way to do it within the rule extraction scenario is by using P@1 rules with respect to the model output. With that, even though they may not be explaining the whole model or are not able to explain every instance, their explanations will be completely aligned with the model.

For feature relevance explanations, the literature shows that they help to see how they contribute to the positive class (outliers in anomaly detection). For rule extraction explanations, they can help to explain outliers with respect to what will turn that outlier into an inlier. Considering this, the explanations will target the inlier class, so the outliers can be explained in a counterfactual approach with respect to the non-anomalous subspace (for local explanations). For global explanations that help to see what feature values are normally associated to outlier situations, the explanations would still target the outlier class.

\subsection{2.4. XAI for OCSVM}
The terms "XAI", "explainable" or "interpretable", together with "OC-SVM" or "OCSVM" only provide 4 results searching them within titles, abstract and/or keywords within Scopus\textsuperscript{\textregistered} One of them is the work of \cite{kauffmann2020towards}. Here, the authors propose a model-specific method based on the concept that OCSVM models can be rewritten as pooling neural networks. Due to the asymmetry between inliers and outliers, they model with a min-pooling over distances for outliers, and a max-pooling over similarities for inliers. Thanks to turning OCSVM models to a neural network, they apply a deep Taylor decomposition (DTD) to obtain explanations in term of input features. DTD serves as a framework to apply layer-wise retropropagation (LRP) in order to obtain the feature contribution of the input features to a predicted output. The authors extend the explanations generated to include using both input features or support vectors. 
In \cite{itani2020one} the authors benchmark different unsupervised ML algorithms for anomaly detection (IsolationForests, OCSVM, Cluster Support Vector Data Description and One-Class decision Tree, OC-Tree), and analyse them over data from the medical domain. They indicate that OC-Tree provides the best results. OC-Tree has the advantage of being a hybrid method that combines the first kernel density estimation for anomaly detection with a decision tree that automatically provides rules that explains the first model. The benchmark of the models is performed in terms of predictive performance, mentioning that OC-Tree is then better for that use case since it directly provides explanations.
In \cite{jang2019anomaly} the authors use OCSVM and Variational Autoencoders for detecting engine faults within 2.4L diesel engines. The faults, which may belong to two types, are precisely the anomalies. For that they use 130 feature parameters. Together with that, they include a post-hoc explainability layer by using LIME (thus, explaining the models in terms of feature relevance).

\cite{padmaja2015hybrid} also shows the combination of OCSVM with XAI. For the XAI part, they use the algorithm Ripper for rule induction. For this algorithm, they use the information from the support vectors from the OCSVM. At the evaluations, they use three different datasets and measure the performance of the rules extracted in terms of Precision, Recall and F1 metrics over the ground truth of the real anomalies. They also train OCSVM models with a RBF kernel.

The previous analysis of the literature shows that even though there are some works regarding XAI and OCSVM, they are either focused in a particular data field, or they do not compare many rule extraction methods (for the previous literature that uses OCSVM with rule extraction). Due to that, there is still an open area regarding the benchmark of rule extraction techniques over OCSVM models for anomaly detection.

\subsection{2.5. Metrics for XAI}

Beyond indicating the importance of both detecting outliers and being able to explain how the decision took place, it is also crucial to quantify the quality of those explanations. There are some recent reviews in the literature that deal with the challenge of providing metrics in XAI, such as \cite{carvalho2019machine}.
In that article, the authors analyse the literature and define a taxonomy of properties that should be considered in the individual explanations generated by XAI techniques. Even though the paper deals with quantifying the quality of the explanations for an individual datapoint, some of them are also applicable for global explanations.
\begin{itemize}
    \item \textbf{Accuracy}: It is related to the usage of the explanations to predict the output using unseen data by the model. 
    \item \textbf{Fidelity}: It refers to how well the explanations approximate the underlying model. The explanations will have high fidelity if their predictions are constantly similar to the ones obtained by the blackbox model. The authors mention how accuracy and fidelity are intertwined: If the explanations have high fidelity (thus, approximate the model well) and the model has high accuracy, the explanations will also have high accuracy. However, the explanations may have high accuracy (because they predict very well over unseen data) while having low fidelity (because they do not approximate well the original model).
    \item \textbf{Consistency}: It refers to the similarity of the explanations obtained over two different models trained over the same input dataset. High consistency appears when the explanations obtained from the two models are similar. However, a low consistency may not be a bad result since the models may be extracting different valid patterns from the same dataset due to the "Rashomon Effect" (seemingly contradictory information is fact telling the same from different perspectives).
    \item \textbf{Stability}: It measures how similar the explanations obtained are for similar datapoints. Opposed to consistency, stability measures the similarity of explanations using the same underlying model.
    \item \textbf{Comprehensibility}: This metric is related to how well a human will understand the explanation. Due to this, it is a very difficult metric to define mathematically, since it is affected by many subjective elements related to human's perception (such as context, background, prior knowledge, etc.). However, there are some objective elements that can be considered in order to measure "comprehensibility", such as whether the explanations are based on the original features (or based on synthetic ones generated after them), the length of the explanations (how many features they include), or the number of explanations generated (i.e. in the case of global explanations). In general terms, using the original features, while keeping the number of explanations generated and the features used to a minimum, will increase comprehensibility.
    \item \textbf{Certainty}: It refers to whether the explanations include the certainty of the model about the prediction or not (i.e. a metric score).
    \item \textbf{Importance}: Some XAI methods that use features for their explanations include a weight associated with the relative importance of each of those features. 
    \item \textbf{Novelty}: Some explanations may include whether the datapoint to be explained comes from a region of the feature space that is far away from the distribution of the training data. This is something important to consider in many cases, since the explanation may not be reliable due to the fact that the datapoint to be explained is very different from the ones used to generate the explanations.
    \item \textbf{Representativeness}: It measures how many instances are covered by the explanation. Explanations can go from explaining a whole model (i.e. weights in linear regression) to only be able to explain one datapoint. 
\end{itemize}

Considering the case of rule extraction techniques, the outputs (rules) for the whole dataset can be analyzed from the perspective of global explanations. In this context, one additional aspect to consider is \textbf{diversity}, a metric that indicates whether the explanations are redundant or repetitive and can already be mostly covered by another explanation, or if they provide insights that are not deducible from the other explanations available. 

From among all these metrics, \cite{barakat2010rule} already commented on the importance of comprehensibility, accuracy and fidelity for rule extraction techniques that explain a SVM model. The metrics are defined as:
\\Accuracy = No. instances classified correctly by the rules / Length test set
\\Fidelity = No. instances where the rule predictions match the model predictions / Length test set
For "consistency", No. of rules and No. of antecedents (analogous to rule size).

\cite{neto2020explainable} analyses XAI metrics for Random Forests by using an Interpretability Matrix that shows the relationship between Rule Coverage - Rule Certainty - Feature Relevance.
\cite{longocomparative} shows a model-agnostic comparative for rule extraction algorithms using C4.5Rule-PANE, REFNE, RxREN and TREPAN. For that, they use 8 datasets of up to 8124 total instances and 40 features. As blackbox models they use Neural Networks models for classification (with different configurations). Finally, they propose several metrics for measuring the quality of the explanations.

\begin{itemize}
    \item \textbf{Completeness:} Percentage of input instances covered by rules over total input instances. Analogous to "Representativeness".
    \item \textbf{Correctness:} Percentage of input instances correctly classified by rules over total input instances. Analogous to "Accuracy".
    \item \textbf{Fidelity:} Percentage of input instances on which the predictions of model and rules agree over total instances.
    \item \textbf{Robustness:} Applying small perturbations over the datapoints that do not change the prediction of the model, the sum of differences between the original prediction and the new prediction, divided by the number of instances analyzed. It is analogous to the concept of "Stability".
    
    \begin{equation}
    \begin{split}
      Robustness = \frac{\sum_{n=1}^{N} f(x_n) - f(x_n + \delta)}{N} \\
    \end{split}
    \end{equation}
    
    Robustness is further analysed in \cite{alvarez2018robustness}, where the authors evaluate it for feature relevance model-agnostic post-hoc XAI techniques (LIME and SHAP).
        
    \item \textbf{Number of rules} and \textbf{Average rule length}, similar to \cite{barakat2010rule}.
\end{itemize}

They apply these metrics and see, using the Friedman's test, that C45-Pane has significantly superior results over all of the datasets considering all of the metrics, followed by TREPAN.
The remaining papers yielded by the query aforementioned either do not deal with metrics for rule extraction techniques, or only focus in the "Accuracy" aspect.

For our research regarding rule extraction, we will focus in analysing the degree of "comprehensibility" of the rules, the coverage of those rules of the datapoints available ("representativeness"), if the rules approximate the underlying model ("stability"), and if they have overlaps among them and are redundant ("diversity"). The advantage of these metrics for unsupervised ML is that they do not need any ground truth information about the "correct" output that the model should have, as opposed to other metrics like "accuracy". This is interesting because many times that ground truth is not available. This is applicable to our use case, since we want to detect anomalies over real industry datasets belonging to Telefónica where there is no prior information about the anomalies. 

The challenge here is defining how to quantify the metrics of stability and diversity (since comprehensibility and representativeness for rule extraction is already defined in the previous SOTA. Stability was also defined previously, but it is important to minimize the number of data points to consider in order to generalize the metric for large datasets (where generating perturbation for every input datapoint would be computationally expensive). This is specially important when working with unsupervised ML models, where many times there is not a reduced test set available, and the evaluations needs to be done using only the training data. Also, when working with P@1 rules, where the fidelity to the original model will always be perfect, it is important to measure the stability in order to see how the explanations behave with unseen data (moreover when there is no test set available). 
Due to that, we will propose and use algorithms to quantify them in a rule extraction scenario applied for unsupervised ML for anomaly detection.

\subsection{2.6. The psychology of explanations}
In ML, explanations are the “key” to open blackbox algorithms, and are therefore likely to play an important role in possible future AI regulations. It is expected that in certain domains or applications, whenever an algorithm takes an autonomous decision or provides a recommendation that has a significant impact on people’s lives, some kind of explanation is required \cite{ec-excellence}. In this paper, we have quantified several quality parameters of explanations generated by different methods for non-supervised learning algorithms, including the comprehensibility of the generated explanations. But for explanations to be comprehensible and effective for people in different situations, we also need to consider the consumer of the explanations: the people.

How do people understand explanations? In psychology, explanations are seen as crucial for human knowledge and learning, as they are considered proofs of understanding. According to Wilkinson \cite{wilkinson2014levels}, there are three kinds of views of explanations: the formal-logical view (an explanation is like a deductive proof given some propositions), the ontological view (events -state of affairs- explain other events), and the pragmatic view (an explanation needs to be understandable by the “demander”). Explanations that are sound from a formal-logical or ontological view, but leave the demander in the dark, are not considered good explanations. For example, a very long chain of logical steps or events (e.g. hundreds) without any additional structure can hardly be considered a good explanation for a person, simply because he or she will lose track. 

Wilkinson introduces two more concepts to define the adequacy of explanations for demanders. The level of explanation refers to whether the explanation is given at a high-level or more detailed level. The right level depends on the knowledge and the need of the demander: he or she may be satisfied with some parts of the explanation happening at the higher level, while other parts need to be at a more detailed level. The kind of explanation refers to notions like causal explanations and mechanistic explanations. Causal explanations provide the causal relationship between events but without explaining how they come about (a kind of “why” question). For instance, smoking causes cancer. A mechanistic explanation would explain the mechanism whereby smoking causes cancer (a kind of “how” question). Causal explanations can be further divided into common-cause explanations (a single cause has several consequences), common-effect explanations (several causes converge to one consequence), and simple linear chain explanations (one causes leads to one consequence) \cite{keil2006explanation}. 

As said, a satisfactory explanation does not exist by itself, but depends on the demander’s need. In the context of ML algorithms, we can distinguish between several typical demanders of explainable algorithms \cite{alej2019explainable}: 

\begin{itemize}
    \item Domain experts: those are the “professional” users of the model, such as medical doctors who have a need to understand the workings of the model before they can accept and use the model.
    \item Regulators, external and internal auditors: like the domain experts, those demanders need to understand the workings of the model in order to certify its compliance with company policies or existing laws and regulations. 
    \item Practitioners: professionals that use the model in the field where they take users’ input and apply the model, and subsequently communicate the result to the users’ situations, such as loan applications. 
    \item Redress authorities: the designated competent authority to verify that an algorithmic decision for a specific case is compliant with the existing laws and regulations.
    \item Users: people to whom the algorithms are applied and that need an explanation of the result. 
    \item Data scientists, developers: technical people who develop or reuse the models and need to understand the inner workings in detail. 
\end{itemize}

In summary, for explainable AI to be effective, the final consumers (people) of the explanations need to be duly considered when designing XAI systems.

\section{3. Our Proposal}
We first describe the intuition behind our rule extraction approach fom an OCSVM model for anomaly detection. Then, we describe in detail the algorithm implementation.

\subsection{3.1. Algorithm Intuition}
We propose using rule extraction techniques within OCSVM models for anomaly detection, by generating hypercubes that encapsulate the non-anomalous data points, and using their vertices as rules that explain when a data point is considered non-anomalous.
As already mentioned in the introduction, \cite{nunez2002rule} proposes an algorithm to extract rules from a SVM model by performing clustering over the datapoints that belong to one of the classes. The clustered datapoints will be used to obtain a geometric surface that enclose the rest of the datapoints inside. There are two ways to accomplish it: building hypercubes or building hyperspheres. This paper will focus the analysis over the first approach: building hypercubes. The paper will also focus in the model-agnostic variant, where the algorithm obtains the furthermost datapoints from inside the cluster as vertices for the hypercube, so they enclose the rest of datapoints of that category inside (the model specific alternative uses the support vectors). In case that the hypercube generated encloses points from the other category, then the number of clusters will be increased, aiming to obtain smaller cubes that could fit the data without including points from the other class. This is done iteratively until no points from the other class are inside the hypercubes, or a maximum number of predefined iterations is reached. During the process, if a hypercube does not contain points from the other class, then that hypercube is translated into a rule, and those datapoints are removed from the following iteration steps.

\begin{figure}[!h]
\centering
  \begin{tabular}{c@{\qquad}c@{\qquad}c}
\includegraphics[width=1\columnwidth]{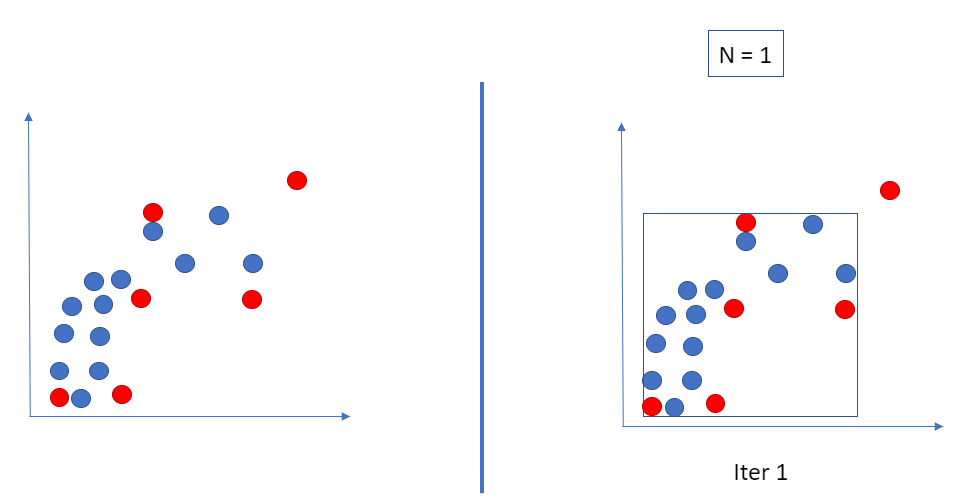}
  \end{tabular} 
  \caption{Clustering over a 2D space. With one cluster over datapoints from one class (blue), there are still others from the other class (red) inside the square.\label{fig:starting_point}}
\end{figure}

\begin{figure}[!h]
\centering
  \begin{tabular}{c@{\qquad}c@{\qquad}c}
\includegraphics[width=1\columnwidth]{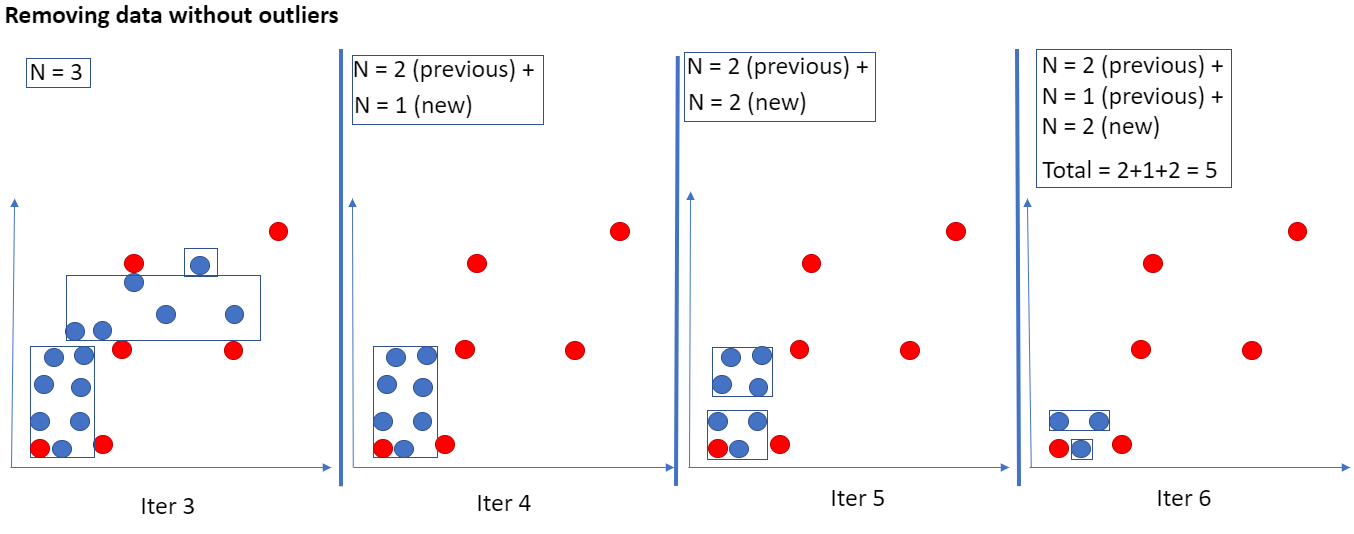}
  \end{tabular} 
  \caption{Applying the proposal of \cite{nunez2002rule}, the number of clusters keeps increasing until no points from the other class are inside, an then that hypercube is translated into a rule.\label{fig:discard}}
\end{figure}

Images \ref{fig:starting_point} and \ref{fig:discard} shows an example application of this algorithm for a 2D space. In Image \ref{fig:starting_point} appears the initial scenario, where the first step in the iteration process consists in applying one cluster over the dataset for datatpoints of one of the classes (blue ones). However, with one cluster, the 2D square that enclose the datapoints contains points from the other class, so more clusters need to be applied. As \ref{fig:discard} shows, iteration 3 (with 3 clusters) is the first one with squares without red points, so those subspaces are turned into rules and the points inside them removed from the iteration process, that starts again with one cluster for the remaining datapoints. Iteration 6 will be the last one, and 5 rules have been extracted up to that point.

\begin{figure}[!h]
\centering
  \begin{tabular}{c@{\qquad}c@{\qquad}c}
\includegraphics[width=1\columnwidth]{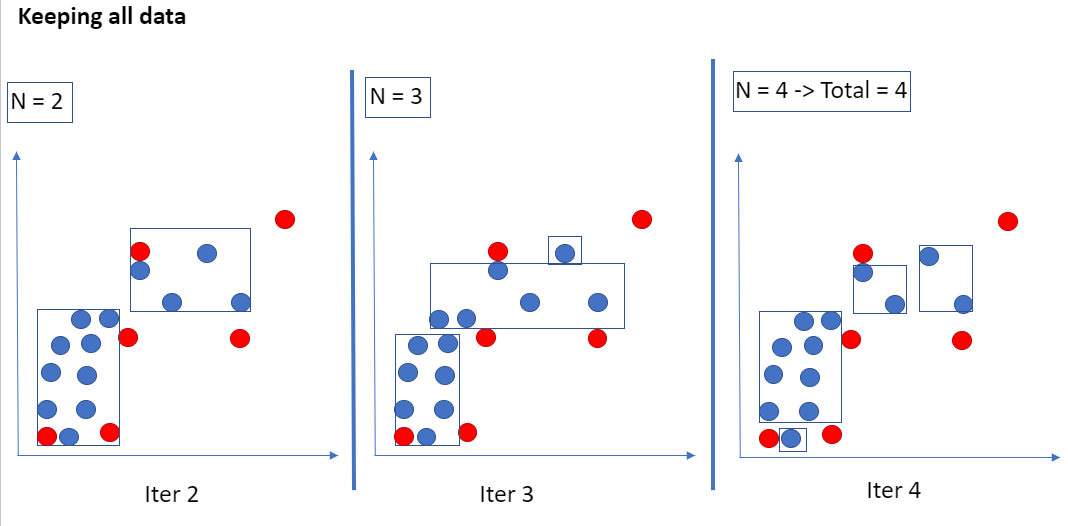}
  \end{tabular} 
  \caption{Keeping all datapoints in every iteration could lead to a reduced number of clusters since there may be data patterns that could only be found in this scenario. \label{fig:keep}}
\end{figure}

The approximation proposed before is not the only one that can be applied in order to extract the rules. Image \ref{fig:keep} shows one of our alternative proposals over \cite{nunez2002rule} method. Instead of removing datapoints that are inside a rule without points from the other class, the process always keeps all datapoints in every iteration since there could be clustering patters that could only be found if all points are together. In this approach, the number of clusters is constantly increased until no datapoints from the other class are inside the hypercubes, or the maximum number of iterations is reached. We will further address this method as "keep" in the remaining of the paper. In contrast, the references to \parencite{nunez2002rule} method will be addressed as "keep\_reset".

\begin{figure}[!h]
\centering
  \begin{tabular}{c@{\qquad}c@{\qquad}c}
\includegraphics[width=1\columnwidth]{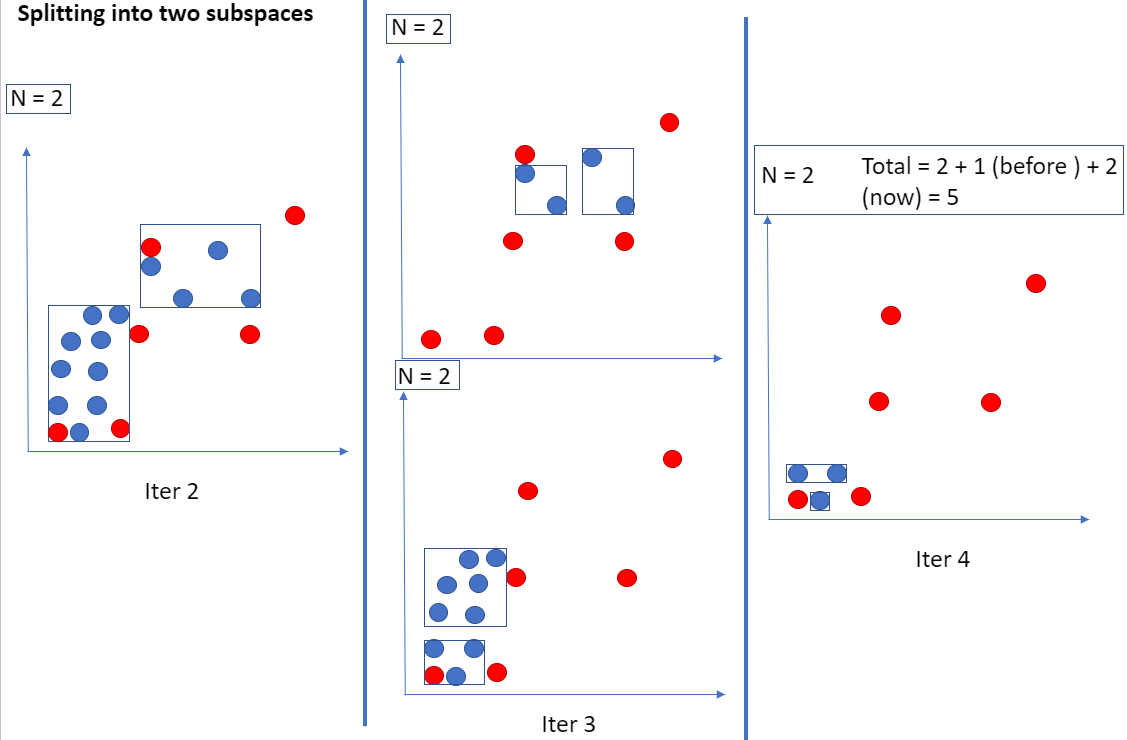}
  \end{tabular} 
  \caption{Splitting subspaces with a binary partition scheme until no red points are inside the rule. \label{fig:keep_reset}}
\end{figure}

Another proposal that we include in this paper over \cite{nunez2002rule} is splitting the subspaces in a binary partition scheme. This is an alternative over the original proposal, that constantly increases the number of clusters until one rule has only datapoints from the same class, and then restarting the clustering process from the beginning for the remaining ones. We will address this method as "split" for the remaining of the paper. Image \ref{fig:keep_reset} shows how the same 2D example using this approach.

According to the taxonomy for XAI in \cite{molnar2019interpretable}, our method has the following characteristics:
\begin{itemize}
    \item Post-hoc: Explainability is achieved using external techniques.
    \item Global and individual: Explanations serve to explain how the whole model works, as well as why a specific data point is considered anomalous or non-anomalous.
    \item Model-agnostic: As with other techniques for global explanations \cite{molnar2019interpretable}, the only information needed to build the explanations are the input features and the outcomes of the system after fitting the model.
    \item Counterfactual: The explanations for why a data point is anomalous also include information on the changes that should take place in the feature values in order to consider that data point as non-anomalous.
\end{itemize}

Since the explanation algorithm is model-agnostic, it can work for any blackbox model. The only information needed is the train dataset and the outputs from the model.  To illustrate it, this paper will show evaluations over OCSVM models with different kernels: radial basis function (RBF) and linear kernel.

Regarding the clustering technique itself, potentially any algorithm could be used, both for \cite{nunez2002rule} or for any of out two proposals over it from this paper. However, there is a caveat that should be considered. The clustering algorithm needs to take into account if the features are only numerical, categorical (non ordinal), or both. 

One algorithm that will be used in this paper for extracting the hypercubes is K-Means ++ \cite{arthur2006k}. However, the standard version of this clustering algorithm is designed for numerical features, and categorical ones should be treated differently. In that case, the approximation would be to extract a rule for each of the possible combinations of categorical values among the data points that are not considered anomalous. Considering again the aforementioned 2-dimensional example, with variable X being binary categorical, a dataset may look like in Figure \ref{fig:outlier5}:

\begin{figure}[h!]
\centering
  \begin{tabular}{c@{\qquad}c@{\qquad}c}
\includegraphics[width=0.70\columnwidth]{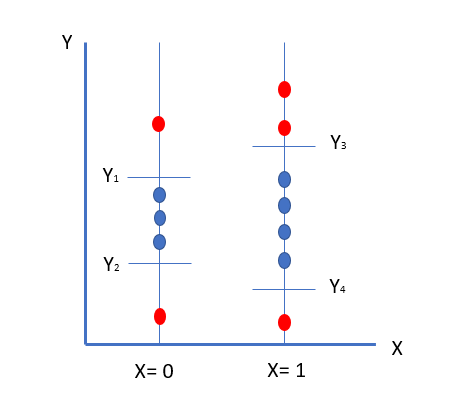}
  \end{tabular} 
  \caption{Rule extraction with a categorical variable.\label{fig:outlier5}}
\end{figure}

In that case, two rules would be extracted, one for each of the possible states of X:
\begin{itemize}
    \item Rule 1: NOT OUTLIER IF X = 0 $\land$ Y $\geq$ Y2 $\land$ Y $\leq$ Y1
    \item Rule 2: NOT OUTLIER IF X = 1 $\land$ Y $\geq$ Y4 $\land$ Y $\leq$ Y3
\end{itemize}

Generally speaking, the algorithm logic can be summarised as:
\begin{itemize}
    \item Apply OCSVM to the dataset to create the model.
    \item Depending on the characteristics of variables, do:
    \begin{itemize}
    \item \underline{Case 1. Numerical only}:  Iteratively create clusters in the non-anomalous data (starting with one cluster) and create a hypercube using the centroid and the points further away from it. Check whether the hypercube contains any data point from the anomalous group; if it does, repeat using one more cluster than before. End when no anomalies are contained in the generated hypercubes. If there are anomalies and the data points in a cluster are inferior to the number of vertices needed for the hypercube, complete the missing vertices with artificial datapoints and end when there are no anomalies or when the convergence criterion is reached.
    \item \underline{Case 2. Categorical only}: The rules will correspond directly to the different value states contained in the dataset of non-anomalous points.
    \item \underline{Case 3. Both numerical and categorical}. This case would be analogous to Case 1, but data points will be filtered for each of the combinations of the categorical variables states. For each combination, there will be a set of rules for the numerical features.
    \end{itemize}
    \item Use these vertices to obtain the boundaries of that hypercube and directly extract rules from them.
\end{itemize}

Besides K-Means++, there are other clustering algorithms that could be applied. In this paper we will analyse also the rules obtained by applying K-Prototypes \cite{ji2013improved}. The advantage of using K-Prototypes is that it can work directly with both categorical and numerical features.

\subsection{3.2. Algorithm Description}
Algorithm \ref{alg:oneclasssvmXAI} contains the proposal for rule extraction for an OCSVM model that may be applied over a dataset with either categorical or numerical variables (or both).
\textit{ocsvm\textunderscore rule\textunderscore extract} is the main function of the algorithm. Regarding input parameters, $X$ is the input data frame with the features, $d_{f}$ a dictionary with two lists ($l_{n}$ a list with the numerical columns and $l_{c}$ a list with the categorical columns), $d_{p}$ is a dictionary with the hyperparameters for  OCSVM (kernel type, upper bound on the fraction of training errors and a lower bound of the fraction of support vectors, $\nu$, and the kernel coefficient, $\gamma$).
This function starts with the feature scaling of the numerical features (function \textit{featureScaling}), followed by the encoding of categorical ones (function \textit{featureEncoding}). After that, it fits an OCSVM model with all the data available and detects the anomalies within it, generating two datasets, $X_y$ with the anomalous data points and $X_n$ with the rest (function \textit{filterAnomalies}). 

The next step is checking the type of features available. If all the features are categorical, then the rules for non-anomalous data points will simply be the unique combination of values for them. If there are both categorical and numerical features, the algorithm obtains the hypercubes (as mentioned for numerical features only) for the subset of data points associated to each combination of categorical values. 

\begin{algorithm}[h!]
\caption{Main pipeline}\label{alg:oneclasssvmXAI}
\begin{algorithmic}[1]
\Procedure{ocsvm\textunderscore rule\textunderscore extract}{$X, d_{f}, d_{p}, m, t$}
    \State $ l_n \gets d_{f}[l_n]$
    \State $ l_c \gets d_{f}[l_c]$
    \State $X[l_n] \gets featureScaling(X[l_n])$
    \State $X \gets featureEncoding(X[l_c])$
    \State $model \gets OneClassSVM(d_{p})$
    \State $model.fit(X)$
    \State $preds \gets model.train(X)$
    \State $distances \gets model.decisionFunction(X)$
    \State $X_y, X_n \gets filterAnomalies(X, preds)$
    \If{$len(l_c)=0$}
        \State $rules \gets getR(X_n, X_y, X, d_{f}, m, t)$
    \ElsIf{$len(l_n)=0$}
        \State $rules \gets getUnique(X_n, l_c)$
    \Else
        \State $cat \gets getUnique(X_n, l_c)$
        \State $rules$ empty list
        \For{$ c\in cat$}
            \State $X_{nf}, X_{yf} \gets filterCat(X_n, X_y, c)$
            \State $rules.append(getR(X_{nf}, X_{ny}, d_{f}, m, t))$
        \EndFor\label{itercategory}
    \EndIf\label{obtainrules}
    \State $rules \gets featureUnscaling(rules, l_n)$
    \State $rules \gets pruneRules(rules, d_{f})$
    \State \textbf{return} $rules$
    \EndProcedure
\end{algorithmic}
\end{algorithm}

Function $getR()$ calls different subfunctions depending on the $t$ parameter value, but in any of the cases, the approach is similar: clustering non-anomalous data points in a set of hypercubes that do not contain any anomalous data points.

The "keep" approach, described in algorithm \ref{alg:oneclass_keep}, iteratively increases the number of clusters (hypercubes) until there are no anomalous points within any hypercube. The function \textit{outPosition} checks whether the rules defined based on the vertices of the hypercube do not include any data point from the anomalous subset, $X_y$. \textit{getRulesKeep} then calls function \textit{getVertex} (described in algorithm \ref{alg:additional})  with a specific number of clusters, $n_{cl}$. This function performs the clustering over the non-anomalous data points, $X_n$, using the function \textit{getClusters} that returns the label of the cluster for each data point, as well as the centroid position for each cluster using the specified cluster algorithm.

If the algorithm is K-Prototypes, then if considers both categorical and numerical features (using $getKP$ function). If is K-Means++, then it applies the clustering over numerical features only (using $getKM$ function).

Then, it iterates through each cluster and obtains the subset of data points for that cluster $X_{nc}$ with the function $insideCluster$. After that, if there are enough data points in that cluster (more data points than the vertices of the hypercube), it computes the distance of each of them to the centroid with \textit{getDist} and uses the furthest $n_v$ as datapoints for obtaining the vertices that enclose the cluster using the \textit{getVertex} function. $n_v$ is a value that represents the hyperspace dimensionality, and is obtained with $hyperDimension$ function.
In case there are less datapoints than the number of vertices that a hypercube of that dimensionality has, then all of them are used for obtaining the vertices. This last scenario does not stop the iterations, since a hypercube in this situation could still include outliers, needing further splitting. As long as there are no outliers inside the rules, they are stored in $rules$ list. However, as soon as there is one rule with outliers inside, then the whole process is repeated again with one more cluster. This keeps taking place until no outliers are inside the rules or the maximum number of iterations is reached.

\begin{algorithm}[h!]
\caption{Rule Extraction - Keeping all datapoints}\label{alg:oneclass_keep}
\begin{algorithmic}[1]
\Procedure{getRulesKeep}{$X_n, X_y,  m, d_f$}
    \State $ l_n \gets d_{f}[l_n]$
    \State $ l_c \gets d_{f}[l_c]$
    \State $max\_iter$ reference value
    \State $check \gets True$
    \State $n_{clusters} \gets 0$
    \While {check}
        \State $rules$ empty list
        \If{$n_{clusters} > max\_iter$} 
            \State $check \gets False$
        \Else
            \State $n_{cl} \gets n_{cl} + 1$
            \State $vInfo \gets getVertex(X_n, X, d_f, m, n_{cl})$
            \For{$iterValue \in vInfo$}
                \State $rules_{cluster} \gets iterValue[0]$
                \State $X_{nc} \gets iterValue[1]$
                \State $l_{y} \gets outPosition(rules_{cluster}, X_y)$
                \If{$len(l_y) = 0$}
                    \State $rules.append(rules_{cluster})$
                    \State $check \gets False$
                \Else
                    \State $check \gets True$
                \EndIf\label{check_anomalies_positions1}
            \EndFor\label{itercluster1}
        \EndIf\label{}
    \EndWhile\label{}
    \State \textbf{return} $rules$
\EndProcedure
\end{algorithmic}
\end{algorithm}

\begin{algorithm}[h!]
\caption{Rule Extraction - Binary partition approach}\label{alg:oneclass_split}
\begin{algorithmic}[1]
\Procedure{getRulesSplit}{$X_n, X_y, m, d_f$}
    \State $ l_n \gets d_{f}[l_n]$
    \State $ l_c \gets d_{f}[l_c]$
    \State $max\_iter$ reference value
    \State $check \gets True$
    \State $l\_sub \gets [X_n]$
    \State $rules$ empty list
    \While {check}
        \If{$len(l\_sub)==0$ or $j > max\_iter$}
            \State $break$
        \EndIf
        \State $l\_{original} \gets l\_sub$
        \State $l\_sub \gets []$ 
        \For{$d$ in $l\_{original}$}
            \State $ n_{cl} \gets 2$
            \State $vInfo \gets getVertex(X_n, X,  d_f, m, n_{cl})$
            \For{$iterValue \in vInfo$}
                \State $rules_{cluster} \gets iterValue[0]$
                \State $X_{nc} \gets iterValue[1]$
                \State $l_{y} \gets outPosition(rules_{cluster}, X_y)$
                \If{$len(l_y) = 0$}
                    \State $rules.append(rules_{cluster})$
                    \State $check \gets False$
                \Else
                    \State $check \gets True$
                    \State $l\_sub \gets l\_sub.append(X_{nc})$
                \EndIf\label{check_anomalies_positions2}
            \EndFor\label{itercluster2}
        \EndFor\label{}
    \EndWhile\label{}
    \State \textbf{return} $rules$
\EndProcedure
\end{algorithmic}
\end{algorithm}

\begin{algorithm}[h!]
\caption{Additional functions}\label{alg:additional}
\begin{algorithmic}[1]
\Procedure{getVertex}{$X_n,  d_f, m, n_{cl}$}
    \State $ l_n \gets d_{f}[l_n]$
    \State $ l_c \gets d_{f}[l_c]$
    \State $n_v \gets hyperDimension(X_n, d_f)$
    \State $d_{bounds}$ empty list
    \State $d_{points}$ empty list
    \If{$m=kprototypes$}
        \State $labels, centroids \gets getKP(X_n, l_n, l_c, n_{cl})$
    \Else
        \State $labels, centroids \gets getKM(X_n, l_n, n_{cl})$
    \EndIf\label{choose_clustering}
    \For{$c\in n_{cl}$}
        \State $X_{nc} \gets insideCluster(labels, X_n)$
        \If{$len(X_{nc}) > n_v$}
            \State $p_{chosen} \gets getDist(X_{nc}, labels[c])$
        \Else
            \State $p_{chosen} \gets X_n$
        \EndIf\label{compute_vertices}
        \State $vertices \gets getVertices(p_{chosen})$
        \State $d_{bounds}.append(vertices)$
        \State $d_{points}.append(X_{nc})$
    \EndFor\label{iter_all_clusters}
    \State \textbf{return} $d_{bounds}, d_{points}$
\EndProcedure
\end{algorithmic}
\end{algorithm}

The "split" approach is defined in algorithm \ref{alg:oneclass_split}. This function has some similarities with \ref{alg:oneclass_keep} with the following differences. Instead increasing the number of clusters in every iteration, $n_{cl}$ is always 2. Also, $l\_sub$ receives the data after every split. Initially, $l\_sub$ contains only one dataset, the inliers $X_n$. However, after another iteration, its value is set to the data from the clusters in which the rules did contain some outlier.

In any of the three methods, after obtaining the rules, function $featureUnscaling$ is used to express rules in their original values (not the scaled ones used for the ML models). And function $pruneRules$ checks whether there are rules that may be included inside others; that is, for each rule it checks whether there is another with a bigger scope that will include it as a subset case.

\subsection{3.3. Influence of the kernel}
As mentioned before, OCSVM models are configured using mainly three hyperparameters: $\nu$, $\gamma$ and the kernel type. Depending on the kernel type, the construction of the decision frontier to differentiate between outliers and inliers changes. In particular, Radial Basis Function (RBF) kernel will find hyperspheres (one or more) that enclose the inliers, leaving outliers outside. 

\begin{figure}[h]
\centering
  \begin{tabular}{c@{\qquad}c@{\qquad}c}
\includegraphics[width=0.60\columnwidth]{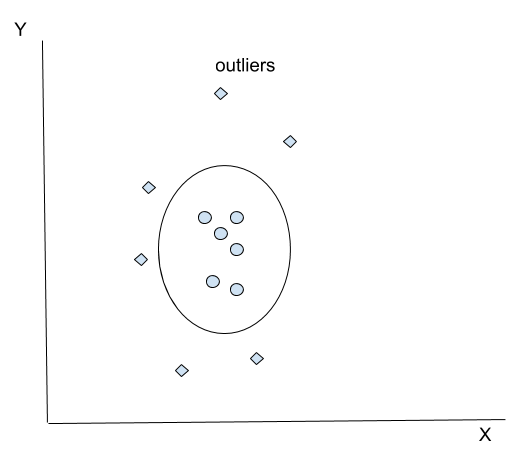}
  \end{tabular} 
  \caption{With an RBF Kernel the correct hypercube will be the one that encloses the points that are not anomalies, since the OCSVM algorithm will try to enclose most of the points inside the decision frontier and leave anomalies outside.\label{fig:outlier4}}
\end{figure}

The diverse density of outliers versus inliers highlights that there may be differences in the rules depending on which class they enclose. Mainly, since the decision function is a hypersphere, the intuition is that it will be easier to find rules that enclose all those points. Figure \ref{fig:outlier4} illustrates this idea.

\subsection{3.4. Algorithms for metrics}
As mentioned before, the metrics considered in this paper are divided into four subsets: comprehensibility, representativeness, stability and diversity. Since, to the best of our knowledge, some of these metrics are not implemented within the main XAI framerworks, we propose within these paper a set of algorithms to compute them in a rule extraction scenario. 
We apply the metrics for the case of unsupervised anomaly detection using OCSVM models, but they could be applied for any model that has a binary output and that is explained through rule-extraction techniques.

\begin{itemize}
    \item \textbf{Metrics for comprehensibility}: Number of rules ($n\_rules$), size of the rules ($size\_rules$).
    \item \textbf{Metrics for representativeness}: Percentage of datapoints explained with P@1 rules ($per\_p1$) and the median percentage coverage of datapoints by each rule ($p1\_coverage$).
    \item  \textbf{Metrics for stability}: How many artificial points (similar to a subset of prototypes from the dataset) are classified by the rules with the same predictions yielded by original blackbox model ($precision\_vs\_model$).
    \item \textbf{Metrics for diversity}: Degree of hyperspace overlapping between all the rules ($score\_intersect$).
\end{itemize}

\subsubsection{Comprehensibility:}
The metrics for "comprehensibility" are directly analyzed from the rules themselves; $n\_rules$ is computed counting the number of rules generated, and $size\_rules$ is computed checking the elements that define the rule (i.e. $X > 3$ AND $X < 7$ AND $Y > 1$ have a $size\_rules=3$ while $X > 3$ have a $size\_rules=1$). This proposal already appears in \cite{barakat2010rule}.

\subsubsection{Representativeness:}
The metric $per\_p1$ for "representativeness" simply checks the percentage of datapoints for the target class explained with P@1 rules. The other metric in this group is $p1\_coverage$. It checks the median performance of the rules themselves: it computes the median percentage of coverage for the target class by each rule. This proposals are similar to \cite{longocomparative}, with the particularity of focusing on P@1 rules.

\subsubsection{Stability:}
The metric $precision\_vs\_model$ computes the "stability" metric of the hypercubes. The first step is obtaining the prototypes from the dataset and generate random samples near them. Then, obtain the prediction of the original model for those artificial samples and checks if the predictions using the rules are the same.

The steps for these metric are described below, and the detailed pseudocode appears in algorithm \ref{alg:stability}. 

Model agreement:
\begin{itemize}
    \setlength{\itemindent}{2em}
    \item Choose N prototypes that represent the original hyperspace of data
    \item Generate M samples close to each of those N prototypes using Protodash algorithm \cite{gurumoorthy2019efficient}; the hypothesis is that close points should be generally predicted belonging to the same class.
    \item For each of those N*M datapoints (M datapoints per each N prototype) check whether the rules (all of them) predict them as inliner or outlier; the datapoints that come into the function are either outliers or inliers. If they are inliers,  then the rules identify an artificial datapoint (of those M*N) as inlier if it is outside every rule. If the datapoints are outliers it's the same reversed: a datapoint is an inlier if no rule includes it.
    \item It then checks if the predictions using the rules for those artificial datapoints are the same as the one provided by the original model.
    \item With that, it computes \% of predictions for the artificial datapoints aforementioned that are the same between the rules and the original OCSVM model.
\end{itemize}

Algorithm \ref{alg:stability} receives the dataset $X$ of inliers/outliers (depending if the rules are computed for inliers or outliers), the rules $X_r$ and the OCSVM fitted and trained model $clf$. Then obtains the protoypes with $ProtodashExplainer()$ function and generates the random samples $X_s$ near them with $randomNear()$, where an upper and lower limits ($th_s$, $th_l$) can be defined for how close are those points to the prototypes. Then, it checks which rules enclose that datapoint with $checkInR()$, and if at least one of them encloses the datapoint, it is considered that it can be classified using the rules. The metric $precision\_vs\_model$ is specified in $n\_precision$ variable, that checks the percentage of agreement between the classifications using the rules and the ones with the model, through $checkInModel()$ function. 

\begin{algorithm}[h!]
\caption{Stability}\label{alg:stability}
\begin{algorithmic}[1]
\Procedure{getAgreement}{$X, X_r, clf$}
    \State $X_p \gets ProtodashExplainer(X)$
    \State $X_s \gets []$
    \For{$p\in X_p$}
        \State $X_s \gets X_s.append(randomNear(p, th_l, th_s))$
    \EndFor\label{generate_samples}
    \State $n\_precision \gets 0$
    \State $l\_rules \gets []$
    \For{$d\in X_s$}
        \State $l\_iter \gets []$
        \For{$r\in X_r$}
             \State $l\_iter \gets l\_iter.append(checkInR(d, r))$
        \EndFor
        \State $r\_rules \gets max(l\_iter)$
        \State $r\_model \gets checkInModel(d, clf)$
        \If{$r\_rules = r\_model$}
            \State $n\_precision \gets n\_precision+1$
        \EndIf\label{compute_agreement}
    \EndFor
    \State $n\_precision \gets n\_precision/len(X_s)$
    \State \textbf{return} $n\_precision$
\EndProcedure
\end{algorithmic}
\end{algorithm}

\subsubsection{Diversity:}
The metric to measure "diversity" is $score\_intersect$, and it analyses if the rules are different with few overlapping concepts. This is computed checking the area of the hypercubes of the rules that overlaps with another one.
The way to check this is by seeing the 2D planes of each hypercube (by keeping two degrees of freedom for the features in the hyperplane coordinates; n-2 features are maintained and the other two are changed between their max/min values in order to obtain the vertices of that 2D plane). Then, it obtains the area of the 2D planes for the rules that overlaps, and each of those 2D areas is turned into a score between 0 and 1 by using the Jaccard similarity index and dividing the area of intersection of the 2D planes by their area of union.

The pseudocode for this metric appears in algorithm \ref{alg:score_intersect}. Algorithm \ref{alg:score_intersect} receives the dataset $X$ of inliers/outliers (depending if the rules are computed for inliers or outliers), the rules $X_r$, the list of columns for numerical features $l\_n$ and the one for categorical $l\_c$.
The first step is obtaining all the two tuples combinations of numerical features, using $combinations()$ function. After that, it obtains the combination of categorical values with function $unique()$. The algorithm then analyses separately the rules that belong to each categorical combination values. For each of those subset of rules $X\_r\_i$, if there are at least two rules, then it defines the tuples of possible rule combinations, $combR$. Then, it iterates per each combination of two numerical features. These two features will correspond to the features that will be changed, leaving the rest of the $l\_fix$ features fixed, in order to extract 2D planes from the hypercubes with $get2D$, and storing those planes in $polys$ variable. Those planes are used for obtaining the Jaccard similarity index with $scorePolys()$ function. If there is an iteration where one of the two dimensions has the same value, it is skipped since the area will be 0. ($checkEqual(pair_f)$).

\begin{figure}[h!]
\centering
  \begin{tabular}{c@{\qquad}c@{\qquad}c}
\includegraphics[width=1 \columnwidth]{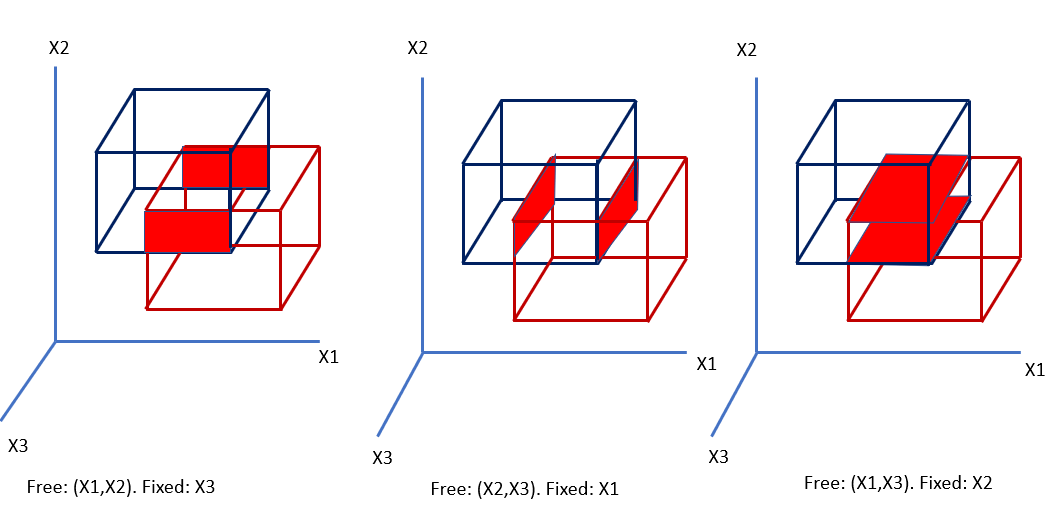}
  \end{tabular} 
  \caption{The overlapping between rules (hypercubes) approximated using their 2D planes' area of intersection.\label{fig:img_cubes}}
\end{figure}

Image \ref{fig:img_cubes} describes the process for an example in a 3D space. Since all the rules translate into a hypercube, we can choose two features at a time (leaving the rest fixed) and obtain the coordinates for those 2D planes (using their vertices values). Then, for two rules, we can see the area of overlapping between those 2D hyperplanes, as well as their area of union. With that areas, we obtain the Jaccard similarity index. Since the Jaccard similarity index ($score\_i$) yields a value between 0 and 1 (0 when there is no overlapping, and 1 when the area of intersection is the same as the area of union in a total overlapping), we can turn it into a metric in order to express a score value by doing $1-score\_i$, so a perfect score will be the one corresponding to no overlap between the rules. This is repeated for all 2D planes of the hypercubes, and we compute the mean of all the individual scores in order to have one final metric ($final\_score$) that is still between 0 and 1, with 1 the perfect score and 0 the worst.

\begin{algorithm}[h!]
\caption{Diversity}\label{alg:score_intersect}
\begin{algorithmic}[1]
\Procedure{getInterScore}{$X, X\_r, l\_n, l\_c$}
    \State $l\_free \gets combinations(l\_n, 2)$
    \State $X\_c \gets unique(X[l\_c])$
    \State $score \gets []$
    \State $n\_inter \gets 0$
    \For{$cat \in rows(X\_c)$}
        \State $X\_r\_i \gets X\_r[cat]$
        \If{$len(X\_r\_i)>2$}
            \State $combR \gets combinations(X\_r\_i, 2)$
            \For{$pair_f \in l\_free$}
                \If{$checkEqual(pair_f)$}
                \State $continue$
                \EndIf
                \State $l\_fix \gets l\_n[!=pair_f]$
                \State $polys \gets get2D(combR, l\_fix,pair_f)$
                \State $score\_i, n\_i \gets scorePolys(polys)$
                \State $score \gets score.appends(1 - score\_i)$
                \State $n\_inter \gets n\_inter + n\_i$
            \EndFor
        \EndIf
    \EndFor
    \State $final\_score \gets mean(score)$
    \State \textbf{return} $final\_score$
\EndProcedure
\end{algorithmic}
\end{algorithm}

All the algorithm that we have proposed for computing XAI metrics are XAI-specific metrics: metrics that are specific for a particular type of XAI technique (in this case, rule extraction).

\subsection{3.5. Pruning rules}
Many of the rules obtained with all the methods described above are suboptimal, since they can be enclosed into another bigger rule. In order to reduce the number of rules, and remove redundancies, we apply a simple pruning technique prior to the computing and evaluation of metrics.
We check every hypercube generated and see if their limits are inside any other rule. If they are, we eliminate that rule from the set of rules. We check this for every rule against every other rule in the dataset, and we keep checking it in a loop until no rules are eliminated.

\subsection{3.6. From local to global rules}
Anchors \cite{ribeiro2018anchors} is one way of extracting rules for XAI. However, Anchors, as mentioned before, is a local method that explains one datapoint with rules. To be able to use it in the evaluation carried out in this paper, we need to turn Anchors into a global method. A simple way to do that is extracting rules for each datapoint of the input dataset, and prune those rules in order to keep the most relevant ones (as done in Section 3.5).
This way, the whole dataset is explained, and the results can be compared with the remaining algorithms.

From the computational cost side, since obtaining Anchors rules for each datapoint is costly, we propose using Protodash \cite{gurumoorthy2019efficient} for scenarios when the dataset is too big. In this case, Protodash will select the relevant prototypes from the dataset, and Anchors will obtain the rules only for those points.

\subsection{3.7. Combining everything}
There is a question that will arise at this point: Which rule would be better? One with better results in "comprehensibility", or one with better results at, for instance, "Diversity"? When there is a need to choose a trade-off, which criteria should be prioritized? 
The answer to this will heavily depend upon the domain needs. However, in general terms, all the metrics can be combined into a single one that offers a unique view over them. It can be done with a metric in terms of $final\_metric = f(C, R, S, D)$ with C representing the comprehensibility metrics, R the representativeness, S the stability and D the diversity. There is another aspect that can be considered while creating a function to encapsulate all metrics. In general, it is better to have a lower value for comprehensibility metrics (less rules, less rule size) since that contributes to an enhancement of comprehensibility. Regarding the rest of the metrics, higher values are better. Thus, a simple way to compute this is adding the results for representativeness, stability and diversity (adjusting their relative importance by a set of weights), and subtracting comprehensibility results. Since the values for the metric of comprehensibility are the only ones that are not in a range of 0 to 1, we scale them before computing this metric in order to have all values in the same range by dividing them with respect to the number of inliers or outliers (number of rules) or by a value based on the number of features (rule size).

With this, a higher final value will be better. This is expressed in Equation \ref{finalmetric}.

\begin{equation}\label{finalmetric}
\begin{split}
  C = {\alpha_1 * (1 - n\_rules) + \alpha_2 * (1 - size\_rules)} \\
  R = \beta_1 * per\_p1 + \beta_2 * p1\_cov \\
  S = \gamma*p\_vs\_m \\
  D = \theta*score\_int \\
  final\_metric = \frac{(R + S + D + C)}{N} \\
\end{split}
\end{equation}

$N$ is equal to 5 in this case since we are considering 5 metrics. The different $\alpha$, $\beta$, $\gamma$ and $theta$ parameters could be adjusted in order to weight the different metrics in case one of them are more important than others. Our proposed methods to compute a general metric is a very naive way to approach it, and more sophisticated ways could be explored. However, its important to highlight the need to be able to analyse everything together for some use cases since there are many XAI aspects to measure and it may difficult the comparison between XAI techniques.

\section{4. Evaluation}
We use our algorithm over different datasets (both public and from Telefonica's real data), to evaluate the following hypotheses:
\begin{itemize}
    \item \textbf{Hypothesis 1 (H1):} The rule extraction method of \cite{nunez2002rule} and our proposed variations applied over OCSVM for anomaly detection using a RBF kernel yield significantly less P@1 rules when applied for explaining inliers than over outliers or when using a linear kernel.
    \item \textbf{Hypothesis 2 (H2):} Our proposed variations over \cite{nunez2002rule} yield similar results for P@1 rules that explains the inliers of an OCSVM anomaly detection model when compared to \cite{nunez2002rule} in terms of explainability regardless of the kernel (considering Linear and RBF).
    \item \textbf{Hypothesis 3 (H3):} The rule extraction method of \cite{nunez2002rule} and our proposed variations yield better results for P@1 rules that explains the inliers of an OCSVM anomaly detection model in terms of explainability than other rule extraction techniques and regardless of the kernel (considering Linear and RBF).
\end{itemize}

As mentioned in Section 3, explanations in terms of rule extraction for anomaly detection may help to see with a counterfactual view what would make an outlier turn into an inlier by explaining the inlier class (for local explanations). For explaining what feature values are normally associated with outlier datapoints (global explanations) these explanations will target the outlier class. This is why \textit{hypothesis 1} checks the contribution of RBF kernel for grouping datapoints inside its hypersphere in order to help explaining them with less rules.

For the hypothesis checks, we will consider the results yielded by the XAI rule extraction methods over different datasets (Section 4.2) together with the type of kernel used for the OCSVM, as well as the type of data points explained (outliers or inliers). Thus, we will have N datasets x 2 types of kernel x 2 types of datapoints. This serves for performing an hypothesis contrast based on the Wilcoxon signed-rank test \cite{conover1998practical}, since it has been proved useful for comparing different ML model metrics results over several datasets for both classification \cite{demvsar2006statistical} and regression tasks \cite{trawinski2012nonparametric}.

\subsection{4.1. Datasets}
The datasets used belong to different domains, have different sizes and different number of features (both categorical and numerical), as indicated in Table \ref{table:1}:
\begin{itemize}
    \item Datasets 1 and 2 are about seismic activity \cite{sathe2016lodes}. Dataset 1 is bi-dimensional with only numerical features ('gdenergy', 'gdpuls'). Dataset 2 has 2 categorical features ('hazard', 'shift') and 7 numerical ('seismoacoustic', 'shift', 'genergy', 'gplus', 'gdenergy', 'gdpuls', 'hazard', 'bumps', 'bumps2').
    \item Dataset 3 is about cardiovascular diseases \cite{padmanabhan2019physician}. There are 4 categorical features ('smoke', 'alco', 'active', 'is\textunderscore man') and 7 numerical ('age', 'height', 'weight', 'ap\textunderscore hi', 'ap\textunderscore lo','cholesterol','gluc').
    \item Dataset 4 is from a call center at Telefónica (TEF Comms). It is real data that includes the total number of calls received in one of its services during every hour. Using these data, some features are extracted (weekday), and they are cyclically transformed, so that each time feature turns into two features for the sine and cosine components. The rules in this case are also transformed back into the original features in order to enhance rule comprehension. 
    \item Dataset 5 contains Telefónica's data about IoT devices attached to cars for vehicle tracking. The data is aggregated in daily windows for each vehicle, representing features that model the daily behaviour of that vehicle. It contains 49 numerical features (such as the number of events with high RPM or the maximum temperature of the coolant), and 12 categorical ones (binary variables that indicate the model and make of that car, among others).
    \item Dataset 6 refers to US census for year 1990 \cite{blake1998uci}. It has 2 categorical features ('dAncstry1\textunderscore 3', 'dAncstry1\textunderscore 4') and 7 numerical ones ('dAge', 'iYearsch', 'iFertil', 'iImmigr', 'iYearwrk', 'dTravtime', 'dRearning').
\end{itemize}

\begin{table}[h!]
\centering
\begin{tabular}{lllll} 
 \textbf{Dataset} & \textbf{Ref.} & \textbf{Nº Cat.} & \textbf{Nº Num.} & \textbf{Nº Rows} \\ [0.5ex]
 \hline
 1 & \cite{sathe2016lodes} & 0 & 2 & 669\\ 
 2 & \cite{sathe2016lodes} & 2 & 7 & 1705\\
 3 & \cite{padmanabhan2019physician} & 4 & 7 & 42000\\
 4 & TEF Comms & 0 & 5 & 2712\\
 5 & TEF Fleet & 12 & 49 & 59844\\
 6 & \cite{blake1998uci} & 2 & 7 & 106819\\  [1ex]
\end{tabular}
\caption{Description of each dataset, with their reference (Ref.), categorical features (Nº Cat.), numerical features (Nº Num) and number of rows.}
\label{table:1}
\end{table}

We ran experiments with the following infrastructure: the implementations of the OCSVM algorithm, the K-Means++ clustering and the DT algorithms are based on Scikit-Learn  \cite{scikit-learn}. The rest of the code described in Algorithms \ref{alg:oneclasssvmXAI} and \ref{alg:oneclass_keep} were developed from scratch, and available in Github \cite{Barbado2019}, together with the algorithms used for measuring the "stability" and the "diversity" of the rules.

\subsection{4.2. Model Configuration}
OCSVM models use as hyperparameters: $\nu = 0.1, \gamma = 0.1, kernel = rbf$ or $\nu = 0.1, \gamma = 0.1, kernel = linear$ for linear kernel. K-means++ models use $max \_ iter = 100, n \_ init = 10, randomState = 0$. K-Prototypes uses $init='Huang', max\_iter=5, n\_init=5$. DT uses default parameters, with $randomState=42$ and Gini criterion to find the best splits. All Protodash applications use $kernelType='Gaussian', sigma=2$, with $m=1000$ for the samples used in the Anchors rule extraction step, and $m=len(rules)$ for the computation of metrics, having m at least a value of 20. 
RuleFit uses $tree\_size=len(feature\_cols)*2$, $rfmode='classify'$ with $len(feature\_cols)$ the number of features that appear in each dataset. RuleFit also considers only rules with a non zero coefficient, and with an importance $>0$. For SkopeRules, since we want only P@1 rules, we use $random\_state=42$, $precision\_min=1.0$, $recall\_min=0.0$. FRL and Anchors use both their default library parameters. BRLG uses $lambda_0=1e-3, lambda_1=1e-3, CNF=False$. LOGRR uses $lambda_0=0.005, lambda_1=0.001, useOrd=True$. GRLM uses $maxSolverIter=2000$ considering only coefficients with value $>0$.

\subsection{4.3. Results}
In this Section we check our hypotheses. We will refer to K-Means approach as KM, and K-Prototypes as KP. Thus, for instance, K-Means with the "split" method will be identified as KM\_split.

\begin{figure}[h!]
\centering
  \begin{tabular}{c@{\qquad}c@{\qquad}c}
\includegraphics[width=1 \columnwidth]{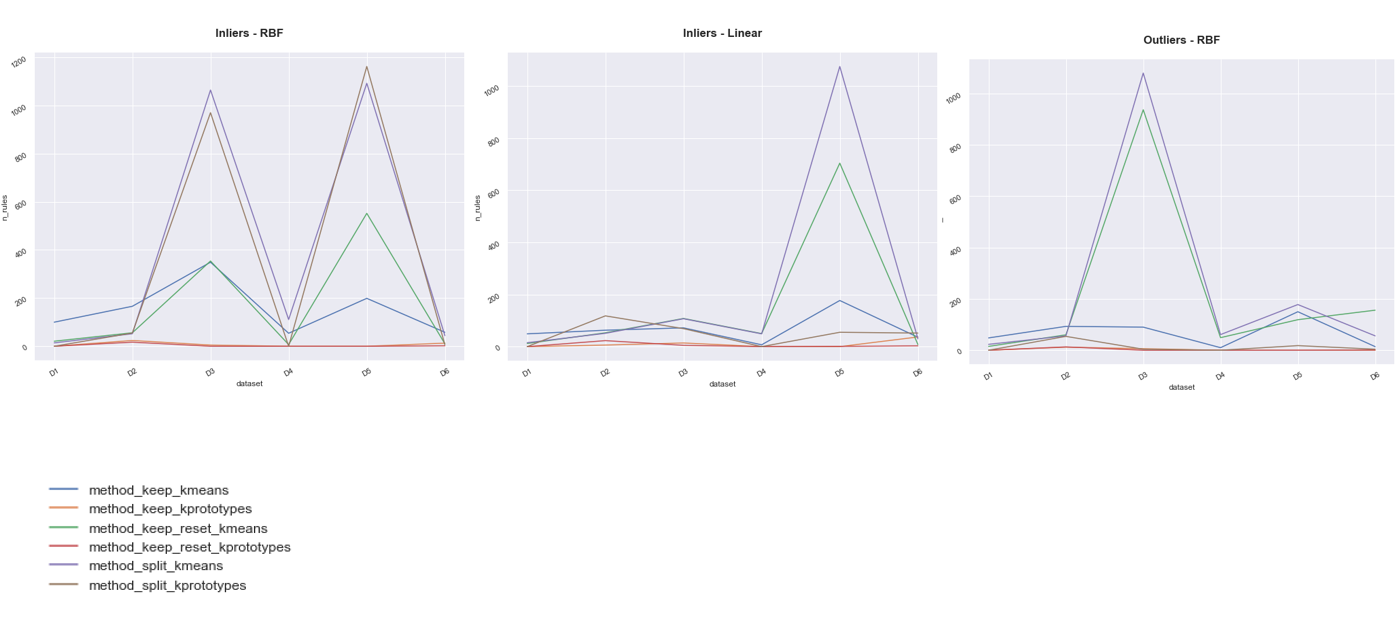}
  \end{tabular} 
  \caption{Number of P@1 rules per clustering-based method and per dataset. K-prototypes methods are not used for D1 and D5 since those datasets do not have categorical columns (there would not be significant differences between the methods). \label{fig:img_n_rules}}
\end{figure}

Figure \ref{fig:img_n_rules} and Table \ref{table:hypothesis1} provide the results associated to \textit{hypothesis 1}. Here, we want to check if there are significantly less P@1 rules for inliers using a RBF kernel, compared to using a linear kernel for inliers, or the same RBF kernel for outliers. For the Wilcoxon signed-rank tests we will compare only combinations of method-kernel-inliers/outliers for datasets that have at least 1 P@1 rule (since, as Figure \ref{fig:img_n_rules} shows, not every method in every dataset is able to yield P@1 rules). Since the comparisons involve few datapoints in some cases, we check against a minimum p-value of 0.1. Considering this, only KM\_split and KM\_keep have significant differences in the number of rules. In those cases, H1 is actually rejected: RBF for inliers yields more rules than either RBF for outliers, or linear for inliers. Regarding the other methods, there are no statistically strong results to conclude anything. This results are shown in Table \ref{table:hypothesis1}. Figure \ref{fig:img_n_rules} shows how at datasets D3 and/or D5 (the ones with more features) the number of rules abruptly increase for some of the methods ("split" ones and KM\_keep\_reset). With that, even though it is not assured for every method, these rule extraction methods when applied to inliers and when using an RBF kernel tend to generate more rules than in the other cases.

After comparing those rule extraction methods in terms of the number of rules in order to see significant differences depending on the type of datapoints (inliers/outliers) and the type of kernel (rbf/linear), we proceed to check H2. Here, we compare the methods considering all the XAI metrics proposed previously. This is done by checking every metric over every combination of dataset, kernel and type of data (inliers/outliers), and performing a Wilcoxon signed-rank test in order to see if there are no significant differences between the methods for each of the metrics. Since the datapoints in this case are superior than those present at H1, we check against a minimum p-value of 0.05. For H2 there is no need to check the size of the rules since they will be the same for all the methods using K-Means and for all the methods using K-Prototypes. We only compare between datasets-kernel-type of data that exists in both methods considered. Thus, the means for the KM methods may vary depending on whether they are compared between them or they are compared against KP ones (and vice versa). 

\begin{figure}[h!]
\centering
  \begin{tabular}{c@{\qquad}c@{\qquad}c}
\includegraphics[width=1 \columnwidth]{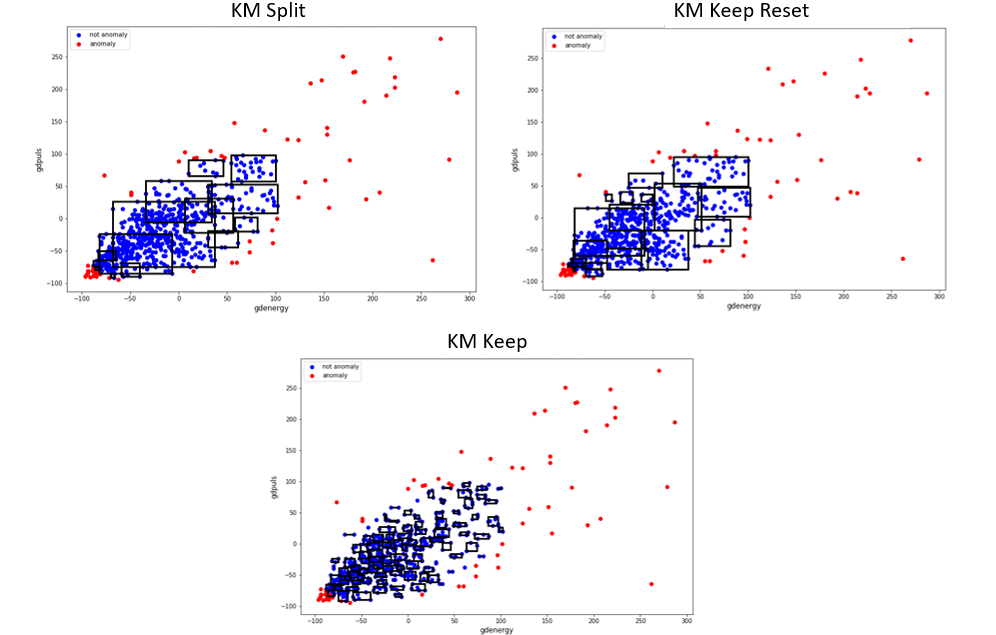}
  \end{tabular} 
  \caption{K-Means based rule extraction methods (for inliers) over D1 dataset with RBF kernel. \label{fig:img_clustering_2D}}
\end{figure}

At Table \ref{table:xai_metrics_h2} we see the methods and metrics that have significant differences according to Wilcoxon signed-rank test. There are some cases where the metrics do differ significantly, as some methods yield better results. This is the case of KM\_split. This method outperforms every other one regarding the percentage of datapoints covered by its P@1 rules (per\_p1). It does so in exchange of yielding a greater number of rules than some of the other methods. Thus, it increases "representativeness" by losing in terms of "comprehensibility". In general, KM methods cover more datapoints with P@1 rules that their counterparts with KP. Considering the other metric from "representativeness", p1\_coverage, there are no significant differences between KM\_split and KM\_keep\_reset, but both methods yield better results than KM\_keep. Thus, usually the P@1 rules that they yield are able to cover more datapoints. This is logical, since the algorithm that yields the rules in the case of KM\_keep tends to generate smaller hypercubes. An example of this can be seen in Figure \ref{fig:img_clustering_2D} for dataset D1. We can see how KM keep indeed yields rules that are smaller than the ones from the other methods.

Regarding "representativeness" (diversity\_score), there are no significant differences between KM methods, but all of them outperform all the KP ones. In terms of "stability", we see no significant difference between any of the methods.
Finally, the general metric (final\_metric), shows that actually KM\_keep outperform KM\_split, and KM\_keep\_reset. Thus, even though KM\_keep had worse results in terms of "representativeness" than the other KM methods, it is compensated by the other metrics.
With this analysis, we see that KM methods appear to be better than KP ones for P@1 rules and for explaining anomalies over a OCSVM model. However, KM methods are more contested; they seem to have similar results in some metrics (KM\_keep\_reset and KM\_discard are very similar between them), while being different in others (mainly compared to KM\_keep in terms of "representativeness"). Thus, H2 is partially supported. 

\begin{figure}[h!]
\centering
  \begin{tabular}{c@{\qquad}c@{\qquad}c}
\includegraphics[width=1 \columnwidth]{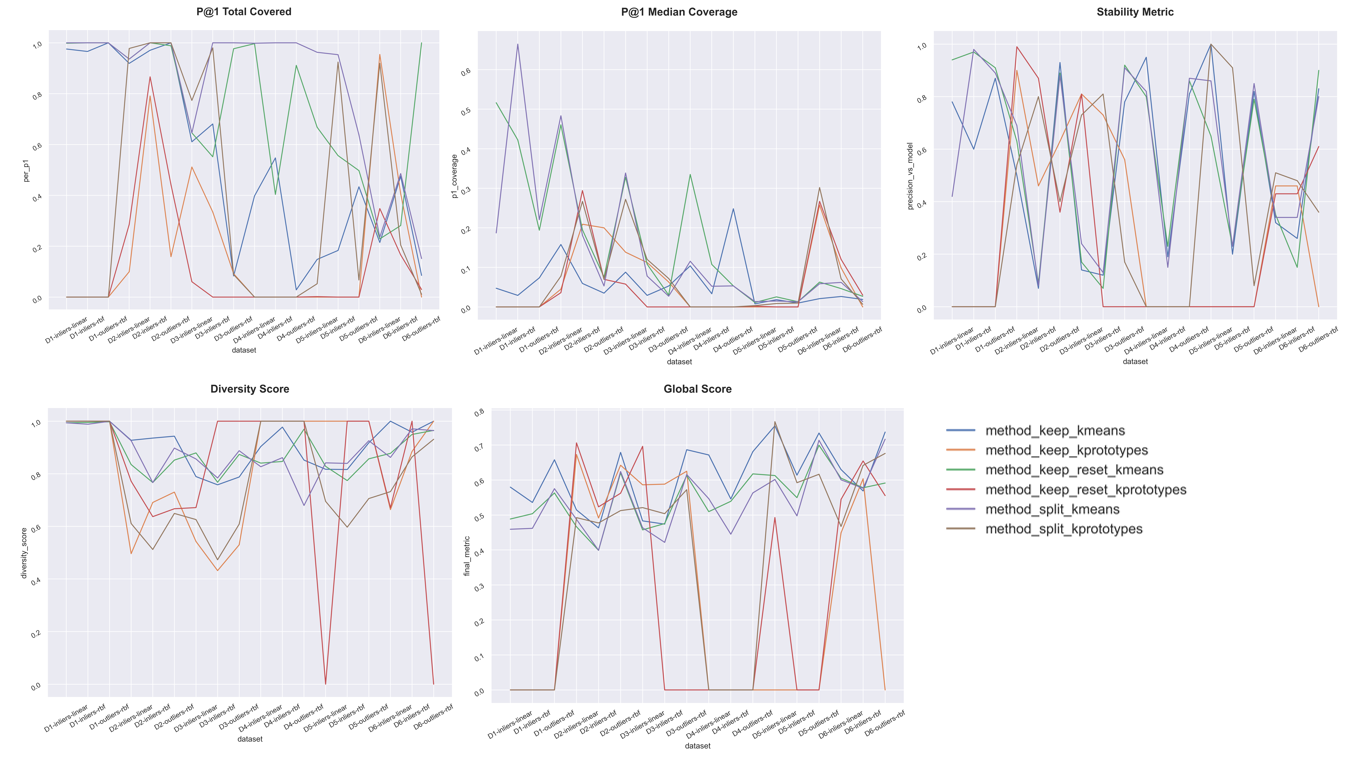}
  \end{tabular} 
  \caption{XAI metrics for the clustering-based rule extraction methods described in this paper. The metrics include representativeness, stability and diversity, as well as a global metric. \label{fig:img_metrics_xai_clustering}}
\end{figure}

A visual analysis for the metrics is provided at
Figure \ref{fig:img_metrics_xai_clustering}. It shows the XAI metrics aforementioned for the clustering-based rule extraction methods and for every combination of dataset, kernel and type of data considered in this paper. 

\begin{figure}[h!]
\centering
  \begin{tabular}{c@{\qquad}c@{\qquad}c}
\includegraphics[width=1 \columnwidth]{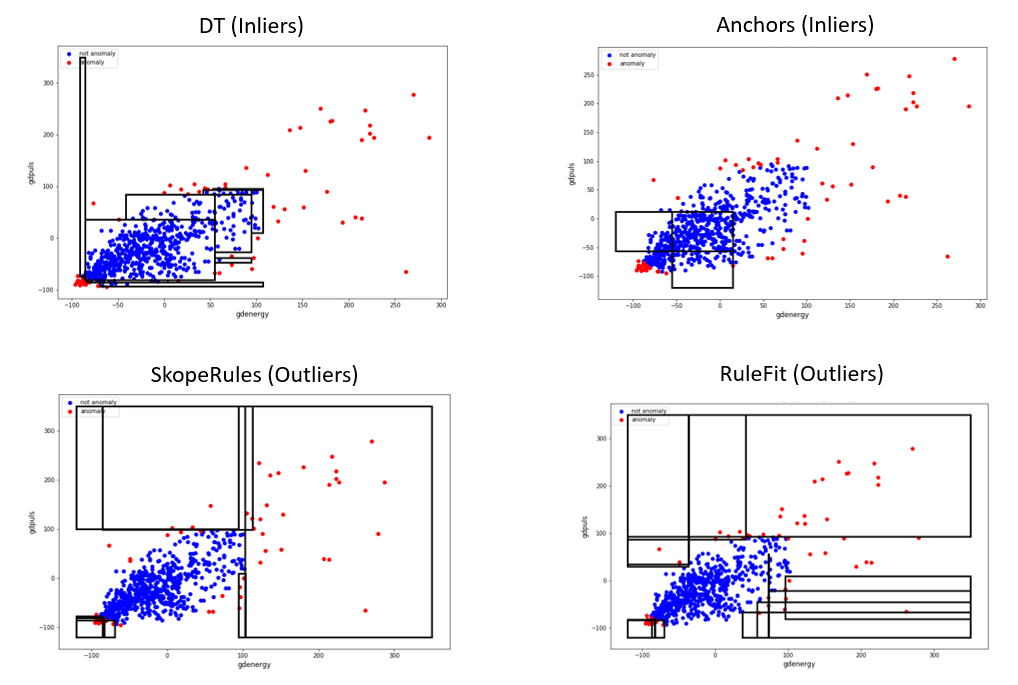}
  \end{tabular} 
  \caption{Visualizations for the rules extracted over D1 with RBF kernel with DT and Anchors (for inliers) and SkopeRules and RuleFit (for outliers). \label{fig:img_other_2D}}
\end{figure}

Finally, we check H3. Since the techniques compared for H2 yield similar results, we will focus only in KM\_split and KM\_keep, and benchmark them against the remaining rule extraction techniques covered in this paper. Figure \ref{fig:img_other_2D} shows visualizations for some of these methods over D1 (when using a RBF kernel).

The results appear in Table \ref{table:xai_metrics-H3}. Here, we see how KM\_split is generally better for every metric except for the ones related to "comprehensibility". In particular, KM\_split is able to significantly cover more datapoints from the target class with P@1 rules (per\_p1) than any of the other methods, and also yields rules that have better coverage (p1\_coverage) than FRL and brlg. However, the mean coverage per rule compared to the other methods is not significantly different. Regarding "stability", KM\_split outperforms brlg, but brlg has a better result in terms of "diversity". KM\_split improves FRL and Anchors in "stability", and DT in "diversity". Finally, considering the general metric, KM\_split has significantly better results than any of the other methods, with the exception of DT, where it does not show significant differences. Considering KM\_keep, the results are similar, as shown in Table \ref{table:xai_metrics-H3}. One difference is that since KM\_keep has a better general metric than KM\_split, it is also able to significantly outperform DT in that aspect. Also, KM\_keep is not outperformed in terms of "diversity" by brlg, as opposed to KM\_split.

As a conclusion, we see that both the solution of \cite{nunez2002rule} with K-Means++ (and with the modification for generating rules for categorical features), together with the variations considered in this paper (for also K-Means++ as clustering method) yield similar results in terms of most of the XAI metrics considered in this paper for explaining the results of a OCSVM anomaly detection model using P@1 rules. The results in terms of comprehensibility (number of rules) are influenced depending on the type of kernel used, and whether they are explaining inliers or outliers. Finally, comparing these techniques with other rule extraction methods, we saw a trade-off between comprehensibility and the remaining XAI metrics. The clustering-bsed rule extraction techniques used in this paper are able to explain better with P@1 rules the results of a OCSVM model (considering the datasets and kernels of this paper) in terms of "representativeness", "stability" and "diversity", but in exchange of "comprehensibility", which is penalized.

\subsection{4.4. Software Used}
The main libraries used for the work done in this paper are the following: 
\begin{itemize}
    \item OCSVM, DT \cite{sklearn-api}
    \item Anchors \cite{alibi} 
    \item Protodash, GRLM, BRLG \cite{aix360-sept-2019}
    \item RuleFit \cite{rulefit}
    \item SkopeRules \cite{skrules}
    \item FRL \cite{bayesian-rules}
\end{itemize}

As we mentioned before, we also provide a repository with all the developed algorithms \cite{Barbado2019}.

\section{5. Conclusion}
In this paper we have analysed the application of XAI techniques over unsupervised outlier detection models through the usage of rule extraction methods applied to OCSVM models. Among the rule extraction techniques, we used both algorithms from the literature, as well as new alternatives that we propose and evaluate together with them. Our first aim was analysing the quality of the rules extracted from a XAI perspective. We have done this by defining metrics for different aspects related to XAI: comprehensibility, representativeness, stability and diversity, as well as proposing a function to aggregate all those metrics together. We evaluated those metrics over different datasets, both from public sources as well as from Telefónica's, using communications and IoT generated data for that purpose. The results for the metrics show that clustering-based techniques yield results that are similar to each other (when usng K-Means++ clustering). When comparing these methods with other rule extraction techniques over different datasets and different kernels, we saw that when working with P@1 rules, the clustering-based methods yielded better results in terms of "representativeness", "stability" and "diversity", in exchange of "comprehensibility", which is penalized by using more rules with more size than other methods.

Our evaluation considered model-agnostic techniques that can be applied over any black-box model. In order to check this empirically, we have used OCSVM models with different types of kernel configurations (linear and RBF). We saw how, indeed, all rule extraction techniques provide similar results regardless of the kernel used.


\subsection{5.1. Limitations of our Approach}
Regarding the XAI metrics and evaluations, the first limitation to consider is that the only model used is OCSVM (with two types of kernel). Even though the algorithms used and the metric definitions are model-agnostic and may be potentially applied over other outlier detection models, the results may differ if other unsupervised models or kernels are used. Together with this, our suggestion of a function that aggregates every metric is a simple baseline that can be further improved.

Also, the analyses carried out are focused in P@1 rules. Thus, the conclusions may be different if all the rules extracted are used, regardless of their precision value. 

Finally, all the rules (for cluster-based methods) and all the checking to see if a data point is inside a hypercube (for all methods) are defined with inequalities ($\leq$, $\geq$). Because of that, the results may be different if we allow values from the other class to be at the limit of the hypercube. 

\subsection{5.2. Future Work}
There are several research lines that can be pursued following the work presented at this paper. The first one to consider is benchmarking these results against other rule extraction techniques that are not covered in this paper. An example is G-Rex algorithms \cite{konig2008g}.
Another research line that can be followed is analysing the metrics of the rule extraction techniques over other unsupervised anomaly detection models, such as IsolationForests \cite{liu2008isolation} or LOF \cite{breunig2000lof}, as well as using other kernel configurations in OCSVM (such as a polynomial one).
Also, while we have proposed a vanilla function to incorporate the metrics belonging to different XAI areas, there is much room of improvement over it in order to find an optimal function that weights appropriately every term.
Finally, rule extraction should also be designed to consider all types of comparisons ($\geq$, $\leq$, $>$ and $<$), and this is something that could also be considered in the cluster-based methods developed.

\nocite{*} 
\printbibliography

\onecolumn
\section{6. Annex}

\begin{table}[H]
\centering
\begin{tabular}{@{}llllll@{}}
\toprule
\multicolumn{1}{c}{\multirow{2}{*}{Method}} & \multicolumn{1}{c}{RBF - Inliers} & \multicolumn{2}{c}{RBF - Outliers} & \multicolumn{2}{c}{Linear - Inliers} \\ \cmidrule(l){2-6} 
\multicolumn{1}{c}{} & mean  & mean  & p             & mean  & p             \\ \midrule
KM\_split            & 396.2 & 242.3 & 1             & 221.2 & \textbf{0.09} \\
KM\_keep\_rest       & 166.7 & 222.7 & 0.56          & 155.8 & 0.84          \\
KM\_keep             & 154.3   & 67.5    & \textbf{0.03} & 67.3  & \textbf{0.03} \\
KP\_split            & 548.3   & 20  & 0.125          & 73.5  & 0.625          \\
KP\_keep\_rest       & 9.5   & 7    & 0.5             & 13    & 0.5             \\
KP\_keep             & 14  & 9     & 1             & 19    & 0.75          \\ \bottomrule
\end{tabular}
\caption{Comparison of the number of rules generated by the different clustering-based rule extraction methods between RBF kernel for inliers, and RBF kernel for outliers or linear kernel for inliers.}
\label{table:hypothesis1}
\end{table}

\begin{table}[h]
\centering
\begin{tabular}{@{}llllll@{}}
\toprule
method 1           & method 2                 & metric           & mean 1 & mean 2 & p-value \\ \midrule

KP\_split          & KP\_keep                 & n\_rules         & 213.92  & 9.75    & 0.0034 \\
KM\_split          & KM\_keep\_reset          & n\_rules         & 286.56  & 181.72  & 0.0342 \\
KM\_keep\_reset    & \textbf{KP\_keep\_reset} & n\_rules         & 259.75  & 5.42    & 0.0005 \\
KM\_split          & KP\_keep                 & n\_rules         & 407.08  & 9.75    & 0.001  \\
KM\_split          & \textbf{KP\_keep\_reset} & n\_rules         & 407.08  & 5.42    & 0.0005 \\
KM\_keep\_reset    & KP\_keep                 & n\_rules         & 259.75  & 9.75    & 0.0024 \\
KM\_keep           & \textbf{KP\_keep\_reset} & n\_rules         & 122.25  & 5.42    & 0.0005 \\
KP\_split          & \textbf{KP\_keep\_reset} & n\_rules         & 213.92  & 5.42    & 0.0005 \\
KM\_keep           & KP\_keep                 & n\_rules         & 122.25  & 9.75    & 0.001  \\
KM\_split          & KM\_keep                 & per\_p1          & 0.83    & 0.54    & 0.0004 \\
KP\_split          & KP\_keep                 & per\_p1          & 0.58    & 0.28    & 0.021  \\
KM\_keep\_reset    & KP\_keep\_reset          & per\_p1          & 0.69    & 0.18    & 0.0015 \\
KM\_keep\_reset    & KP\_keep                 & per\_p1          & 0.69    & 0.28    & 0.0161 \\
KM\_keep           & KP\_keep\_reset          & per\_p1          & 0.48    & 0.18    & 0.0034 \\
KM\_keep           & KP\_keep                 & per\_p1          & 0.48    & 0.28    & 0.0269 \\
KP\_split          & KP\_keep\_reset          & per\_p1          & 0.58    & 0.18    & 0.001  \\
KM\_keep           & KM\_keep\_reset          & per\_p1          & 0.54    & 0.76    & 0.0312 \\
\textbf{KM\_split} & KP\_keep\_reset          & per\_p1          & 0.75    & 0.18    & 0.001  \\
\textbf{KM\_split} & KP\_keep                 & per\_p1          & 0.75    & 0.28    & 0.0093 \\
\textbf{KM\_split} & KM\_keep\_reset          & per\_p1          & 0.83    & 0.76    & 0.0231 \\
KM\_keep           & \textbf{KM\_keep\_reset} & p1\_coverage     & 0.06    & 0.17    & 0.0023 \\
KM\_split          & KM\_keep                 & p1\_coverage     & 0.15    & 0.06    & 0.0104 \\
\textbf{KM\_keep}  & KP\_keep                 & diversity\_score & 0.89    & 0.75    & 0.0329 \\
KM\_keep\_reset    & KP\_split                & diversity\_score & 0.85    & 0.67    & 0.0005 \\
KM\_split          & KP\_split                & diversity\_score & 0.88    & 0.67    & 0.0005 \\
\textbf{KM\_keep}  & KP\_split                & diversity\_score & 0.89    & 0.67    & 0.0005 \\
KM\_split          & \textbf{KM\_keep}        & final\_metric    & 0.54    & 0.61    & 0.0    \\
\textbf{KM\_keep}  & KM\_keep\_reset          & final\_metric    & 0.61    & 0.55    & 0.0001 \\
\bottomrule
\end{tabular}
\caption{Wilcoxon signed-rank hypothesis contrast for the methods and metrics where there are significant differences.}
\label{table:xai_metrics_h2}
\end{table}

\begin{table}[h]
\centering
\begin{tabular}{@{}llllll@{}}
\toprule
method 1           & method 2      & metric               & mean 1 & mean 2 & p-values \\ \midrule
KM\_split          & DT            & n\_rules             & 286.56  & 7.72    & 0.0    \\
KM\_split          & FRL           & n\_rules             & 286.56  & 2.17    & 0.0    \\
KM\_split          & SkopeRules    & n\_rules             & 286.56  & 2.83    & 0.0    \\
KM\_split          & RuleFit       & n\_rules             & 286.56  & 21.44   & 0.0028 \\
KM\_split          & Anchors       & n\_rules             & 286.56  & 20.44   & 0.0    \\
KM\_split          & \textbf{brlg} & n\_rules             & 286.56  & 0.22    & 0.0    \\
KM\_split          & \textbf{brlg} & size\_rules          & 29.0    & 0.22    & 0.0    \\
KM\_split          & RuleFit       & size\_rules          & 29.0    & 3.33    & 0.0    \\
KM\_split          & DT            & size\_rules          & 29.0    & 5.72    & 0.0003 \\
KM\_split          & Anchors       & size\_rules          & 29.0    & 3.39    & 0.0    \\
KM\_split          & SkopeRules    & size\_rules          & 29.0    & 1.69    & 0.0    \\
KM\_split          & FRL           & size\_rules          & 29.0    & 2.17    & 0.0    \\
\textbf{KM\_split} & DT            & per\_p1              & 0.83    & 0.28    & 0.0008 \\
\textbf{KM\_split} & Anchors       & per\_p1              & 0.83    & 0.14    & 0.0    \\
\textbf{KM\_split} & SkopeRules    & per\_p1              & 0.83    & 0.22    & 0.0001 \\
\textbf{KM\_split} & RuleFit       & per\_p1              & 0.83    & 0.45    & 0.0031 \\
\textbf{KM\_split} & FRL           & per\_p1              & 0.83    & 0.12    & 0.0    \\
\textbf{KM\_split} & brlg          & per\_p1              & 0.83    & 0.02    & 0.0    \\
\textbf{KM\_split} & FRL           & p1\_coverage         & 0.15    & 0.08    & 0.0442 \\
\textbf{KM\_split} & brlg          & p1\_coverage         & 0.15    & 0.02    & 0.0007 \\
\textbf{KM\_split} & FRL           & precision\_vs\_model & 0.58    & 0.26    & 0.004  \\
\textbf{KM\_split} & brlg          & precision\_vs\_model & 0.58    & 0.04    & 0.0    \\
\textbf{KM\_split} & Anchors       & precision\_vs\_model & 0.58    & 0.35    & 0.0268 \\
\textbf{KM\_split} & DT            & diversity\_score     & 0.88    & 0.59    & 0.0432 \\
KM\_split          & \textbf{brlg} & diversity\_score     & 0.88    & 0.89    & 0.0268 \\
\textbf{KM\_split} & RuleFit       & final\_metric        & 0.54    & 0.27    & 0.0005 \\
\textbf{KM\_split} & SkopeRules    & final\_metric        & 0.54    & 0.33    & 0.0003 \\
\textbf{KM\_split} & FRL           & final\_metric        & 0.54    & 0.21    & 0.0003 \\
\textbf{KM\_split} & Anchors       & final\_metric        & 0.54    & 0.26    & 0.0007 \\
\textbf{KM\_split} & brlg          & final\_metric        & 0.54    & 0.04    & 0.0 \\
\bottomrule
\end{tabular}
\caption{Wilcoxon signed-rank hypothesis contrast for the methods and metrics where there are significant differences, comparing KM\_split method against the remaining rule-extraction techniques covered in this paper.}
\label{table:xai_metrics-H3}
\end{table}

\begin{table}[h]
\centering
\begin{tabular}{@{}llllll@{}}
\toprule
method 1          & method 2      & metric               & mean 1 & mean 2 & p-value \\ \midrule
KM\_keep          & Anchors       & n\_rules             & 122.25  & 29.25   & 0.0005 \\
KM\_keep          & RuleFit       & n\_rules             & 122.25  & 18.08   & 0.0015 \\
KM\_keep          & \textbf{brlg} & n\_rules             & 122.25  & 0.22    & 0.0005 \\
KM\_keep          & SkopeRules    & n\_rules             & 122.25  & 3.42    & 0.0005 \\
KM\_keep          & DT            & n\_rules             & 122.25  & 3.0     & 0.0005 \\
KM\_keep          & FRL           & n\_rules             & 122.25  & 2.08    & 0.0005 \\
KM\_keep          & RuleFit       & size\_rules          & 40.0    & 2.75    & 0.0005 \\
KM\_keep          & Anchors       & size\_rules          & 40.0    & 3.92    & 0.0005 \\
KM\_keep          & FRL           & size\_rules          & 40.0    & 1.67    & 0.0005 \\
KM\_keep          & DT            & size\_rules          & 40.0    & 5.25    & 0.0005 \\
KM\_keep          & SkopeRules    & size\_rules          & 40.0    & 2.12    & 0.0005 \\
KM\_keep          & \textbf{brlg} & size\_rules          & 40.0    & 0.22    & 0.0005 \\
\textbf{KM\_keep} & Anchors       & per\_p1              & 0.48    & 0.03    & 0.0005 \\
\textbf{KM\_keep} & DT            & per\_p1              & 0.48    & 0.03    & 0.001  \\
\textbf{KM\_keep} & brlg          & per\_p1              & 0.48    & 0.02    & 0.0005 \\
\textbf{KM\_keep} & FRL           & per\_p1              & 0.48    & 0.1     & 0.0005 \\
\textbf{KM\_keep} & SkopeRules    & per\_p1              & 0.48    & 0.19    & 0.021  \\
\textbf{KM\_keep} & brlg          & p1\_coverage         & 0.04    & 0.02    & 0.0005 \\
\textbf{KM\_keep} & Anchors       & p1\_coverage         & 0.04    & 0.0     & 0.0033 \\
\textbf{KM\_keep} & FRL           & precision\_vs\_model & 0.5     & 0.22    & 0.0425 \\
\textbf{KM\_keep} & brlg          & precision\_vs\_model & 0.5     & 0.04    & 0.0005 \\
\textbf{KM\_keep} & DT            & diversity\_score     & 0.89    & 0.43    & 0.0068 \\
KM\_keep          & \textbf{brlg} & diversity\_score     & 0.89    & 1.0     & 0.0051 \\
\textbf{KM\_keep} & FRL           & final\_metric        & 0.61    & 0.18    & 0.001  \\
\textbf{KM\_keep} & SkopeRules    & final\_metric        & 0.61    & 0.43    & 0.0015 \\
\textbf{KM\_keep} & RuleFit       & final\_metric        & 0.61    & 0.16    & 0.0005 \\
\textbf{KM\_keep} & DT            & final\_metric        & 0.61    & 0.35    & 0.0269 \\
\textbf{KM\_keep} & Anchors       & final\_metric        & 0.61    & 0.21    & 0.001  \\
\textbf{KM\_keep} & brlg          & final\_metric        & 0.61    & 0.0     & 0.0005 \\ \bottomrule
\end{tabular}
\caption{Wilcoxon signed-rank hypothesis contrast for the methods and metrics where there are significant differences, comparing KM\_keep method against the remaining rule-extraction techniques covered in this paper.}
\label{table:xai_metrics-H3}
\end{table}

\end{document}